%% file: main.tex
\documentclass{article}

\usepackage{microtype}
\usepackage{graphicx}
\usepackage{subfigure}
\usepackage{booktabs} 
\usepackage{tabularx}
\usepackage{stfloats}
\usepackage{threeparttablex}
\usepackage{enumitem}
\usepackage{xcolor}
\usepackage[hyphens]{url}
\usepackage{xurl} 
\usepackage{hyperref}
\usepackage{eso-pic}

\usepackage{cleveref}
\crefname{figure}{Figure}{Figures}
\Crefname{figure}{Figure}{Figures}
\crefname{table}{Table}{Tables}
\Crefname{table}{Table}{Tables}

\newcommand{\summarybox}[1]{%
\begin{tcolorbox}[
    colback=gray!10!white,
    colframe=black,
    arc=0mm,
    boxsep=1pt,       
    left=2pt,         
    right=2pt,        
    top=2pt,          
    bottom=2pt,       
    before skip=4pt,  
    after skip=4pt    
]
    \textbf{Summary:}~#1
\end{tcolorbox}
}

\usepackage[accepted]{mlsys2025}

\mlsystitlerunning{Breaking the Ice: Analyzing Cold Start Latency in vLLM}

\usepackage{cleveref,tcolorbox}
\usepackage{todonotes}

\begin{document}

\twocolumn[
\mlsystitle{Breaking the Ice: Analyzing Cold Start Latency in vLLM}



\mlsyssetsymbol{equal}{*}

\begin{mlsysauthorlist}
\mlsysauthor{Huzaifa Shaaban Kabakibo}{de}
\mlsysauthor{Animesh Trivedi}{ibm}
\mlsysauthor{Lin Wang}{de}
\end{mlsysauthorlist}

\mlsysaffiliation{de}{Paderborn University, Paderborn, Germany}
\mlsysaffiliation{ibm}{IBM Research Europe, Zurich, Switzerland}

\mlsyscorrespondingauthor{Lin Wang}{lin.wang@uni-paderborn.de}

\mlsyskeywords{Machine Learning, MLSys, vLLM, Cold Start Latency, LLM Inference, Startup Time Prediction, Serverless Scheduling
}

\vskip 0.3in

\input{src/abstract}

]



\printAffiliationsAndNotice{}  

\newcommand{\badgeshift}{2cm}

\AddToShipoutPictureFG*{%
  \put(\LenToUnit{\paperwidth-2.5cm},\LenToUnit{\paperheight-2.1cm}){%
    \makebox[0pt][r]{%
      \href{https://www.acm.org/publications/policies/artifact-review-and-badging-current}{%
        \includegraphics[width=1.7cm]{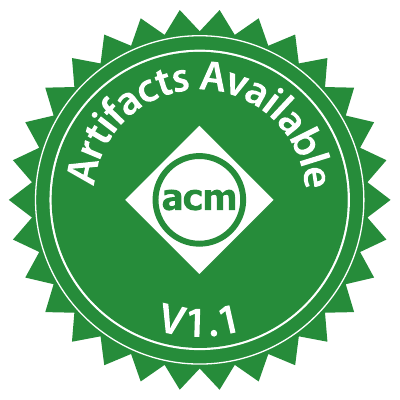}%
      }%
      \hspace{0.3cm}%
      \href{https://www.acm.org/publications/policies/artifact-review-and-badging-current}{%
        \includegraphics[width=1.7cm]{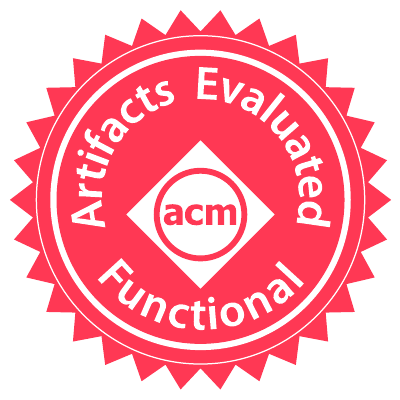}%
      }%
      \hspace{0.3cm}%
      \href{https://www.acm.org/publications/policies/artifact-review-and-badging-current}{%
        \includegraphics[width=1.7cm]{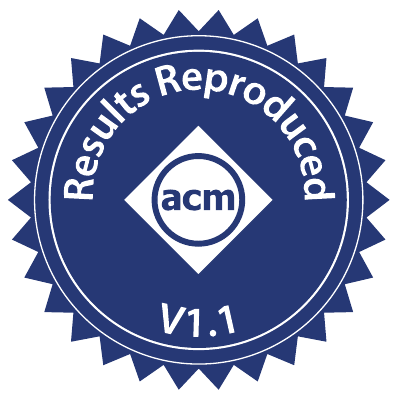}%
      }%
    }%
  }%
}

\input{src/intro_motivation}
\input{src/experiments}
\input{src/impact}
\input{src/modeling2}
\input{src/new-discussion}
\input{src/related-work}
\input{src/conclusion}

\newpage

\bibliography{example_paper}
\bibliographystyle{mlsys2025}
\textbf{Notes:} IBM is a trademark of International Business Machines Corporation, registered in many jurisdictions worldwide. Intel and Intel Xeon are trademarks or registered trademarks of Intel Corporation or its subsidiaries in the United States and other countries. Linux is a registered trademark of Linus Torvalds in the United States, other countries, or both. Java and all Java-based trademarks and logos are trademarks or registered trademarks of Oracle and/or its affiliates. Other products and service names might be trademarks of IBM or other companies.

\appendix
\include{src/appendix}


\end{document}

%% file: src/abstract.tex
\begin{abstract}

As scalable inference services become popular, the cold start latency of an inference engine becomes important. Today, vLLM has evolved into the de-facto inference engine of choice for many inference workloads. 
Although popular, due to its complexity and rapid evolution, there has not been a systematic study on the startup latency of its engine.
With major architectural innovations under it (e.g., the \texttt{V1} API, introduction of \texttt{torch.compile}), in this paper, we present the first detailed performance characterization of vLLM startup latency. We break down the startup process into six foundational steps and demonstrate that this process is predominantly CPU-bound. Each step exhibits consistent and interpretable scaling trends with respect to model- and system-level parameters, enabling fine-grained attribution of latency sources.
Building on these insights, we develop a lightweight analytical model that accurately predicts vLLM's startup latency for a given hardware configuration, providing actionable guidance for resource planning in large-scale inference environments. All our benchmarking datasets, analysis tools, and prediction scripts are open-sourced at \href{https://github.com/upb-cn/vllm-startup-profiler}{https://github.com/upb-cn/vllm-startup-profiler}.

\end{abstract}

%% file: src/intro_motivation.tex
\section{Introduction}\label{sec:introduction}

Despite the success of Large Language Models (LLMs) in various 
domains~\cite{llm-usecases1, llm-usecases2, llm-usecases3}, deploying LLMs 
at scale still poses significant challenges with respect to GPU resource provisioning, 
request scheduling, and performance scaling~\cite{2025-nsdi-superserve}. 
To address these issues, serverless computing has emerged as an attractive paradigm for LLM serving~\cite{fu2024serverlessllm, hu2025deepflow, paraserve, qin2024mooncake, zeng2025medusa}. In this paradigm, users provide LLMs while the serverless platform dynamically provisions resources to match workload variations, enabling a pay-as-you-go model that enhances cost efficiency by scaling automatically on-demand. Despite these advantages, serverless deployments face a critical challenge: \textbf{cold start latency}. Under bursty workloads, cloud providers frequently spin up new cold LLM container instances to handle traffic spikes, introducing significantly higher latency, often orders of magnitude greater than serving requests on warm instances~\cite{zeng2025medusa, du2020catalyzer, oakes2018sock}. This latency primarily impacts the Time-to-First-Token (TTFT), a key performance metric in LLM inference~\cite{agrawal2024taming, zeng2025medusa, fu2024serverlessllm}.

A growing body of work has sought to mitigate the cold start latency through techniques such as accelerated checkpoint loading~\cite{fu2024serverlessllm}, reducing runtime initialization overhead~\cite{akkus2018sand, fuerst2021faascache, li2022help, roy2022icebreaker, yu2024rainbowcake}, fast state materialization~\cite{zeng2025medusa}, and pipeline parallelism~\cite{paraserve}.
However, these efforts focus on individual components of the startup process, with limited analysis of the process as a whole. This gap hinders our ability to design scalable and efficient serverless systems to meet the performance demands of LLM inference requests.

This gap is particularly evident in \textbf{vLLM}~\cite{vllm}, a widely adopted and rapidly evolving open-source framework for LLM inference. Despite its widespread adoption, the startup behavior of vLLM still lacks a clear structural understanding within the community. This is reflected in multiple user discussions and issue reports that attempt to diagnose or mitigate startup latency, often without a shared decomposition of the underlying steps~\cite{discussion_1,discussion_2,discussion_3,discussion_4,discussion_5}. Only recently has a dedicated startup-time benchmark been added to the vLLM codebase, suggesting that this aspect of the system had not yet been systematically characterized~\cite{vllm_benchmark}. In this work, we provide the first detailed study of vLLM startup process. Specifically, \textit{our goal is to characterize this process end-to-end, identifying the key steps, quantifying their performance dependencies, and analyzing their GPU-, CPU- and I/O-dependencies.} We argue that answering these questions is both challenging and timely for three key reasons: 

\begin{figure}[!t]
    \centering
    \includegraphics[width=0.45\textwidth]{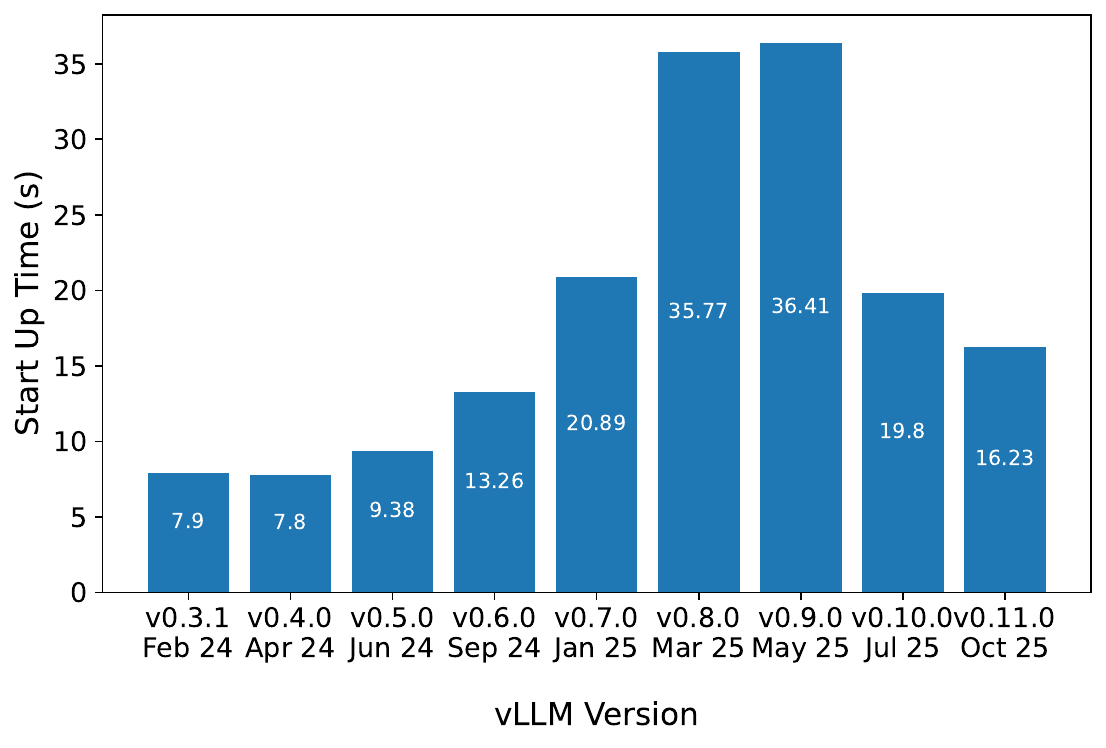}
    \vspace{-0.3cm}
    \caption{Startup times of different vLLM versions using the OPT-6.7B model on an H100 GPU (lower is better).}
    \label{fig:vllm-version-startuptime}
    \vspace{-0.6cm}
\end{figure}

\textbf{Firstly, the popularity and complexity of vLLM.} Over the past few years, vLLM has rapidly become one of the most widely used inference engines, evolving quickly through frequent, community-driven releases~\cite{vllm,vllm-popular}. It delivers a highly optimized inference path~\cite{vllm-anatomy} with techniques such as PagedAttention and prefix caching~\cite{vllm}, chunked prefill~\cite{agrawal2023sarathi}, disaggregated prefill-decoding~\cite{zhong2024distserve} and continuous batching~\cite{yu2022orca}.  
However, this fast evolution also complicates cold start optimizations, as prior techniques often become obsolete due to API changes or with the emergence of new features (if not forward ported) like the \texttt{V1} API and \texttt{torch.compile}~\cite{vllm_deprecate_v0_2025,vllm_v1_torch_compile,vllm_v1_torch_compile2}.
Furthermore, with a growing codebase ($\sim$280K lines of Python code as of \texttt{v0.10}) and new integrations, it has become increasingly more difficult to systematically assess how changes to individual components affect the overall cold start performance~\cite{vllm_startup_time_heavy_hitters_2025}.
To quantify the impact of this complexity, in ~\cref{fig:vllm-version-startuptime} we show the vLLM startup latencies for the last nine major releases during the past 1.5 years. As evident from the figure, there is more than 4$\times$ variance in latencies, and a $2\times$ latencies reduction observed between \texttt{v0.9} and \texttt{v0.10}. These results indicate that there is a need systematically characterize and understand the startup process of vLLM. 

\textbf{Secondly, heterogeneous inference ecosystem.} Modern inference deployments are heterogeneous, where a variety of hardware (CPUs, GPUs, and storage devices) and software (framework, ecosystem, workload, models) parameters can influence resource management efficiency---an important factor in serverless computing. From a hardware point of view, vLLM supports a wide variety of AI accelerators with more than a dozen reported in active usage in 2024~\cite{vllm-usage-trace}. In terms of software, different versions of vLLM are used in the wild. We perform an analysis using \textit{pepy.tech} website~\cite{pepy2025} that reports \texttt{pip install} usage of various Python packages, including vLLM. Over the past three months, all vLLM versions shown in~\cref{fig:vllm-version-startuptime} have been actively installed, ranging from \texttt{v0.3} ($\sim$30K downloads) to \texttt{v0.10} (around 1.3 million downloads). 
Moreover, serving systems commonly host multiple models with diverse sizes, families, architectures (e.g., transformers, mixture-of-experts, and hybrid designs), and popularity~\cite{vllm-usage,yu2025prism}.
Furthermore, in terms of resource management, LLM requests remain highly bursty in nature, making provisioning for the peak demand both costly and inefficient. We analyze multiple publicly available LLM production traces (Microsoft Azure~\cite{cortez2017resource}, Shanghai AI Lab~\cite{hu2024characterization}, Mooncake AI~\cite{qin2024mooncake}, and Alibaba~\cite{chen2025gyges} traces) and report peak-to-mean ratios of 2-20$\times$, revealing significant variance in request arrival rates, thus making resource provisioning challenging. Similar numbers were also reported in the past literature~\cite{2025-nsdi-superserve}. The diversity in hardware, software, models, and workloads underscores the importance of developing a systematic understanding of the startup process to ensure efficient resource management across this heterogeneous ecosystem.

\textbf{Lastly, emergence of containerized, scalable distributed inference blueprints.} 
Since early 2025, LLM ecosystem has seen a rapid emergence of end-to-end, containerized frameworks designed to support distributed and scalable inference. These frameworks provide full blueprints for running production-level LLM services with tightly integrated components, such as an inference request router, a KVCache manager, and worker autoscalers. Examples include NVIDIA Dynamo~\cite{nvidia_dynamo}, Red Hat LLM-D~\cite{redhat_llmd}, AIBrix~\cite{aibrix2025}, and vLLM Production Stack~\cite{vllm_stack2025}.  
A key requirement shared across these frameworks is the ability to scale GPU workers efficiently, which in turn requires an accurate model of the startup cost for each worker instance~\cite{workload_variant_autoscaler}. 
For example, NVIDIA Dynamo recommends performing multi-hour offline profiling of inference environments, followed by periodic online monitoring to build detailed performance models for efficient resource provisioning~\cite{nvidia_dynamo_load_planer}. Such profiling captures end-to-end inference behavior, including model execution, scheduling, and runtime dynamics, to inform autoscaling and workload placement decisions. However, these system-level models typically treat startup as part of a larger inference lifecycle rather than isolating it as a distinct component of container initialization~\cite{aws2025dynamo}.
Hence, an accurate and detailed startup characterization is an essential contribution for designing impactful autoscaler policies for containerized, serverless inference frameworks.  

With these trends, in this paper, we take a step back and focus on developing a comprehensive and systematic understanding of vLLM's startup process.  
We focus specifically on the vLLM engine initialization phase, while intentionally controlling for distributed factors such as container startup, remote storage, and network effects, in order to isolate the core system behavior.
We decompose the startup process into six foundational steps and identify that the overall process is largely CPU-bounded. By examining the scaling characteristics of each step, we uncover consistent and interpretable relationships among model configuration, system environment, and startup latency. Leveraging these insights, we develop a lightweight analytical model capable of predicting vLLM startup time for a given hardware and model configurations. This predictive model enables more informed scheduling and autoscaling decisions in serverless deployments, allowing cloud platforms to plan resource allocation and mitigate cold starts effectively.

%% file: src/experiments.tex
\section{Performance Characterization of vLLM Startup Process}
\label{sec:experiments}

\begin{figure}[t]
    \centering
    \includegraphics[width=0.85\linewidth]{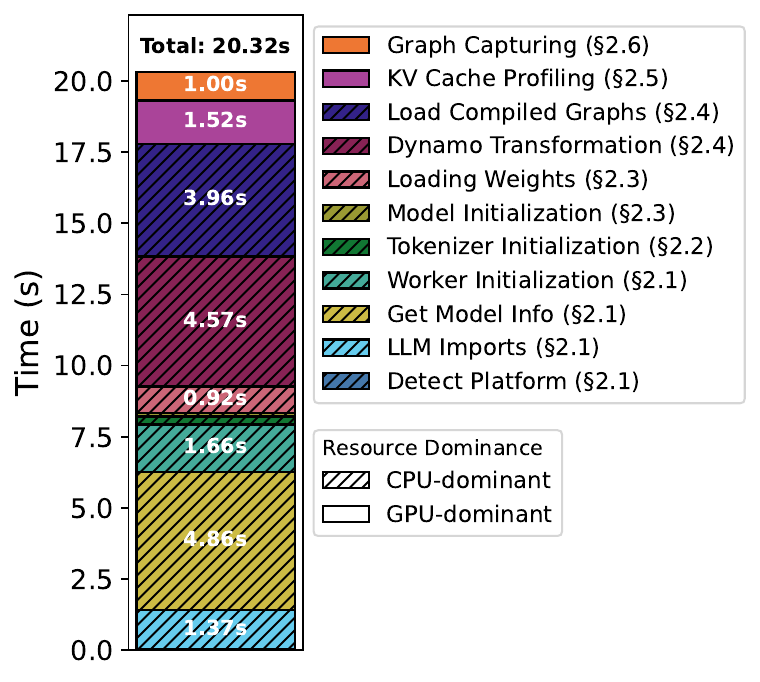}
    \vspace{-0.3cm}
    \caption{vLLM startup latency breakdown with Llama3.2-3B.}
    \label{fig:vllm-init}
    \vspace{-0.5cm}
\end{figure}

\input{src/tables/model-configs}
\input{src/tables/system-env}

We start by conducting a detailed performance characterization of vLLM's startup process. Throughout this paper, we define the startup time as the duration between the start of the inference engine’s initialization and the point at which the engine becomes fully operational and ready to serve inference requests (e.g., when vLLM prints \textit{Application startup complete}). Our goal in performance characterization is to (i) identify unique steps involved in the startup process; (ii) synthesize their performance and scaling dependencies on vLLM-internal configuration and external factors. We perform our analysis with 22 LLMs, detailed in ~\cref{tab:model-configs}, with a variety of architectures and configurations on NVIDIA H100 (\S\ref{sec:experiments}) and L40S (\S\ref{sec:env-config}) GPUs~\cite{nvidia_h100, nvidia_l40s} using vLLM \texttt{v0.10.1.1}. More details of our system environments are shown in~\cref{tab:system-env}. Unless mentioned otherwise, all experiments are conducted on node \texttt{n1} equipped with H100 GPU and AMD EPYC 9354 CPU. For clarity and space reasons, models are selectively omitted in certain plots without loss of generality. To ensure visual consistency, each model is represented using a fixed color across all figures, following the mapping illustrated in~\cref{tab:model-configs}. All experiments are conducted five times, and the reported results represent the average time.

Our analysis reveals six unique steps in vLLM's startup process as shown in~\cref{fig:vllm-init} for Llama3.2-3B~\cite{meta-llama3.2-3b}. 
The figure also shows the dominant resource type for each step and its contributions towards startup latency of 20.32~secs. As shown, all steps are CPU-bound, except for the final two steps (e.g., KVCache profiling and CUDA graph capture), which are GPU-bound, thus establishing that the overall startup process is predominantly bounded by CPU. In the following subsections, we provide a detailed explanation of each step, covering both, their functional role in the startup process, and their potential performance implications. We believe that the upcoming vLLM releases will optimize these steps instead of eliminating them, thus ensuring longevity of this analysis and insights produced.   

\subsection{vLLM Framework Bootstrapping}
\label{sec:bootstrap}

The first step of vLLM’s startup process is Framework Bootstrap, which configures the runtime environment before any model components are loaded. In this step, vLLM initializes the runtime environment and launches an OpenAI-compatible inference API server responsible for managing inference requests. It comprises four substeps:

\textbf{Detect Platform.} vLLM probes the available hardware backend by importing runtime modules (e.g., CUDA or CPU) and confirming device support~\citep{CUDA_Guide_2025}. This determines the appropriate execution environment for subsequent stages.

\textbf{Dependencies Imports.} After backend detection, the framework loads its core dependencies, such as PyTorch, Transformers, tokenizer libraries, and vLLM-specific plugins. The dynamic import and symbol resolution of these packages can introduce noticeable latency of several seconds.

\textbf{Get Model Info.} The API server retrieves the model’s configuration and tokenizer metadata from a local repository or via remote querying (e.g., HuggingFace). This step involves file or network I/O and JSON parsing of files such as \texttt{config.json} and \texttt{model\_index.json}, which define the model architecture and supported tasks.

\textbf{Worker Initialization.} Finally, vLLM spawns its main worker via Ray or Python multiprocessing, setting up inter-process communication, shared memory, and GPU contexts to prepare the runtime for model loading and inference.

Overall, the Framework Bootstrap step is mainly governed by vLLM’s internal implementation and shows stable latency within the same environment across all models, regardless of their parameters. Recent optimizations, such as caching the \texttt{Model\_Class} metadata after the first run~\cite{vllm_pr_cache_model_info}, have reduced the latency of the \textit{Get Model Info} substep from roughly 4.47~secs, to about 0.12~secs. Since this feature was introduced in~\texttt{v0.11}, it was not enabled in our experiments, which covered releases up to~\texttt{v0.10.1.1}.

\summarybox{vLLM framework bootstrapping latency depends mainly on its implementation, and is independent of the model used. Users have little direct control over this step, though on-going optimizations are reducing its impact in newer vLLM versions.}

\begin{figure}[t!]
    \centering
    \includegraphics[width=0.98\linewidth]{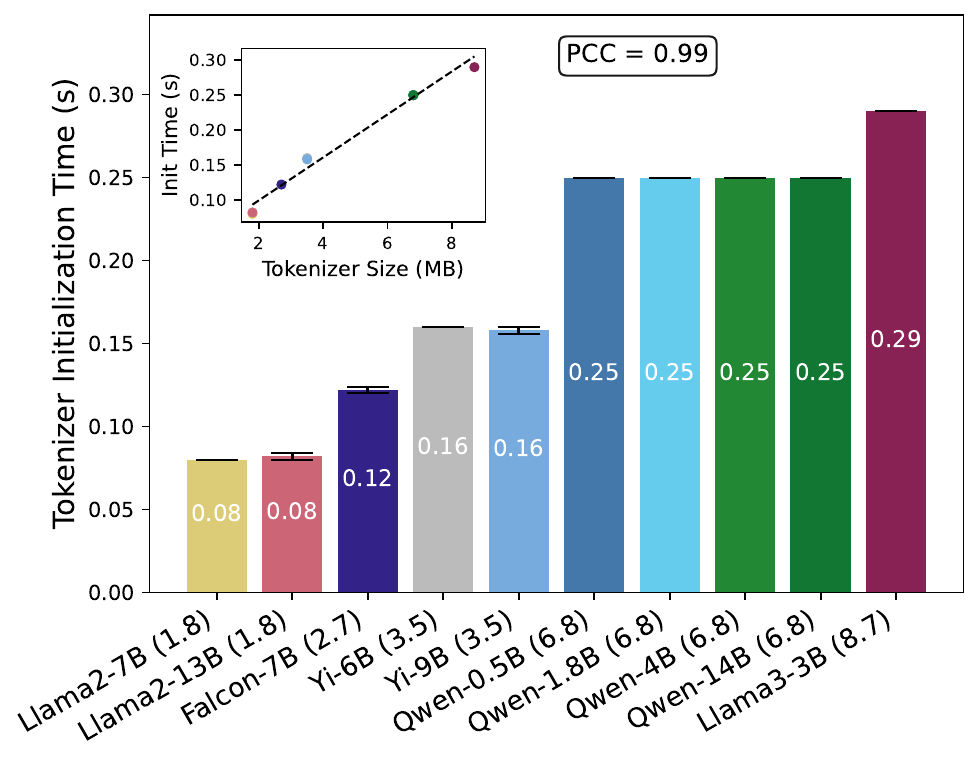}
    \vspace{-0.3cm}
    \caption{Strong linear relationship between the tokenizer size (shown in parentheses) and tokenizer initialization time.}
    \label{fig:tokenizer-init}
    \vspace{-0.5cm}
\end{figure}

\subsection{Tokenizer Initialization}
\label{sec:tokenizer}

The second step of vLLM’s startup process is tokenizer initialization, which prepares the component that converts raw text into a numerical form that the model can process~\cite{singh2024tokenizationcountsimpacttokenization}. This step can be divided into two parts: (i) loading the tokenizer’s vocabulary files and configurations, and (ii) mapping user input text into model-compatible token IDs. However, during the startup process, only the first part occurs where vLLM loads the tokenizer's vocabulary and builds the internal logic required for tokenization. 

The bar chart in~\cref{fig:tokenizer-init} shows the tokenizer initialization time across models. We observe that models like Llama2-7B and Falcon-7B initialize much faster (0.08~secs), while models with larger tokenizers, such as Llama3-3B, take longer (0.29~secs). Additionally, the Qwen series, all sharing the same tokenizer size (6.8~MB), exhibit identical initialization time (0.25~secs), regardless of the model size.

The inset plot in~\cref{fig:tokenizer-init} shows a strong linear relationship between tokenizer size and initialization time, confirmed by a regression fit with a Pearson correlation coefficient (PCC) of 0.99. PCC is a measure ranging from -1.0 (no corelation) to +1.0 (perfect corelation) indicating the strength of a linear relationship between two variables. This shows that initialization latency is primarily governed by the size of tokenizer files, which contain the vocabulary, merge rules, and metadata. As detailed in~\cref{tab:model-configs}, tokenizer size mainly depends on vocabulary size, with minor effects from tokenizer type and encoding format (e.g., JSON vs.~binary). Although different models employ distinct tokenization methods (e.g., SentencePiece for LLaMA, BPE for Falcon), the tokenizer type itself has little impact; overall, tokenizer’s file size dominates initialization performance.

\summarybox{Tokenizer initialization time scales linearly with the tokenizer size, which in turn is determined by the model vocabulary size.}

\subsection{Model Loading}
\label{sec:model-loading}

The third step is model loading. It occurs in two steps: initializing the model structure and loading the pretrained weights. An LLM consists of two main components: its architecture (e.g., layers, activation functions, and attention blocks) and its pretrained weights (i.e., the parameters that define how the model performs its tasks). These components are usually distributed in separate files, which are read and loaded into the GPU memory during the model loading step. 

\textbf{Initializing The Model Structure}.
In this step, vLLM creates the model architecture in memory, setting up layers, attention blocks, and activation functions as defined by the model's configuration file. We identify this step to be independent of model parameter size and other model parameters (e.g., hidden size, FFN dimension).
In our experiments, this step consistently takes about $0.1 \pm 0.05$ s across models, regardless of their size or architecture.

\begin{figure}[t]
    \centering
    \includegraphics[width=0.88\linewidth]{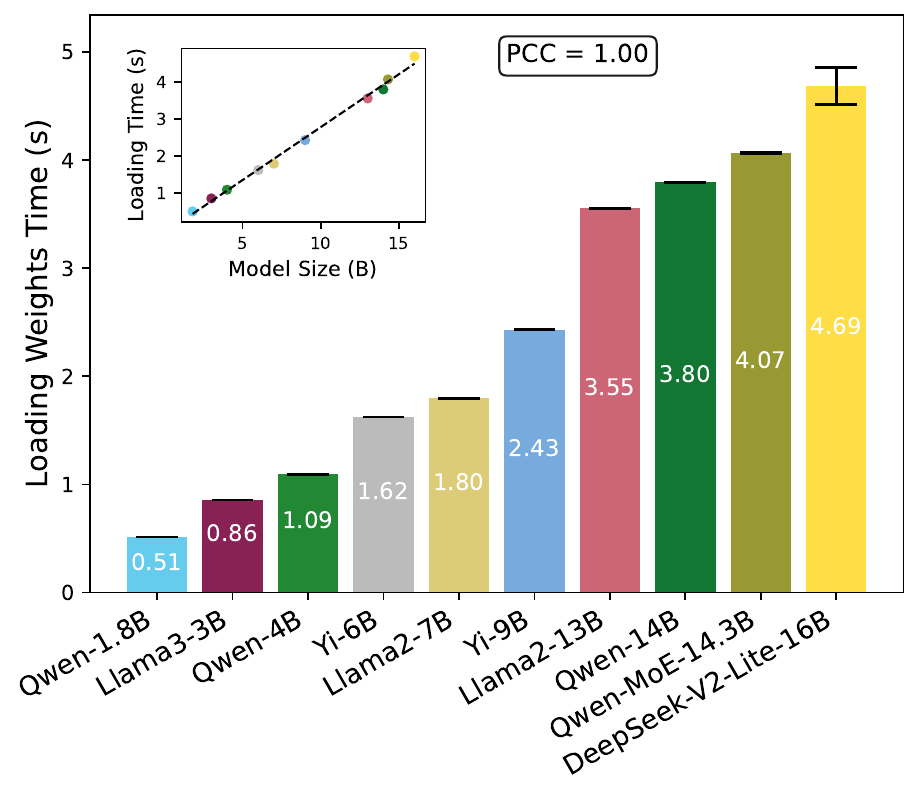}
    \vspace{-0.3cm}
    \caption{Strong linear relationship between the model size and loading weights time.}
    \label{fig:load-weights}
    \vspace{-0.5cm}
\end{figure}

\textbf{Loading The Weights}.
This step involves transferring the pretrained parameters (the weights for the attention block, FFN, etc.) from checkpoint files to GPU memory. 
\cref{fig:load-weights} shows the latency of this step across a variety of models loaded with \texttt{FP16} format. The size of the model depends primarily on the number of model parameters and the numeric precision used to store them. In the figure, the parameter count is reflected in the model names (e.g., 1.8B denotes 1.8 billion parameters). For example, a model such as Qwen-1.8B contains approximately 1.8 billion parameters; when stored in \texttt{FP16} format (2 bytes per parameter), the model requires about 3.6~GB of data to be loaded into GPU memory. In the bar chart, we see that models like Qwen-1.8B and Llama3-3B, with smaller parameter counts, have relatively low loading times around 0.5–1~secs, while larger models such as DeepSeek-V2-Lite-16B take longer, reaching nearly five seconds. Note that these experiments are done with a warm Linux buffer cache, i.e., model checkpoint files are effectively read from the system DRAM. We quantify the impact of SSD loading in~\S\ref{sec:impact-ssd}.

The loading times in \cref{fig:load-weights} exhibit a clear trend: larger models require more time to load their weights, in direct proportion to their size. 
The inset plot further supports this observation by showing a linear relationship between the model size and loading time, with PCC = 1 and a regression line confirming that the loading time increases predictably as model size grows. 
Nonetheless, secondary factors such as quantization or precision format, may influence this step by reducing data volume and transfer time.

\summarybox{Model loading is dominated by the weight-loading time, which \textit{linearly} depends on the model parameter size and the numeric precision.}

\subsection{Torch Compilation}
\label{sec:torch-compilation}

The fourth step in vLLM’s startup process is the \texttt{torch.compile} step, introduced in version~\texttt{v0.7.0} as a major optimization milestone. It leverages PyTorch’s compilation infrastructure to convert Python-level execution into optimized, low-level kernels, reducing Python overhead and enabling kernel fusion for faster inference~\cite{torch-compile-overview}. In vLLM, this process includes two substeps: (i) \emph{Dynamo Bytecode Transformation} and (ii) \emph{Loading/Storing Compiled Graphs}~\cite{torch-compile-integration}.

\begin{figure}[t]
    \centering
    \includegraphics[width=0.98\linewidth]{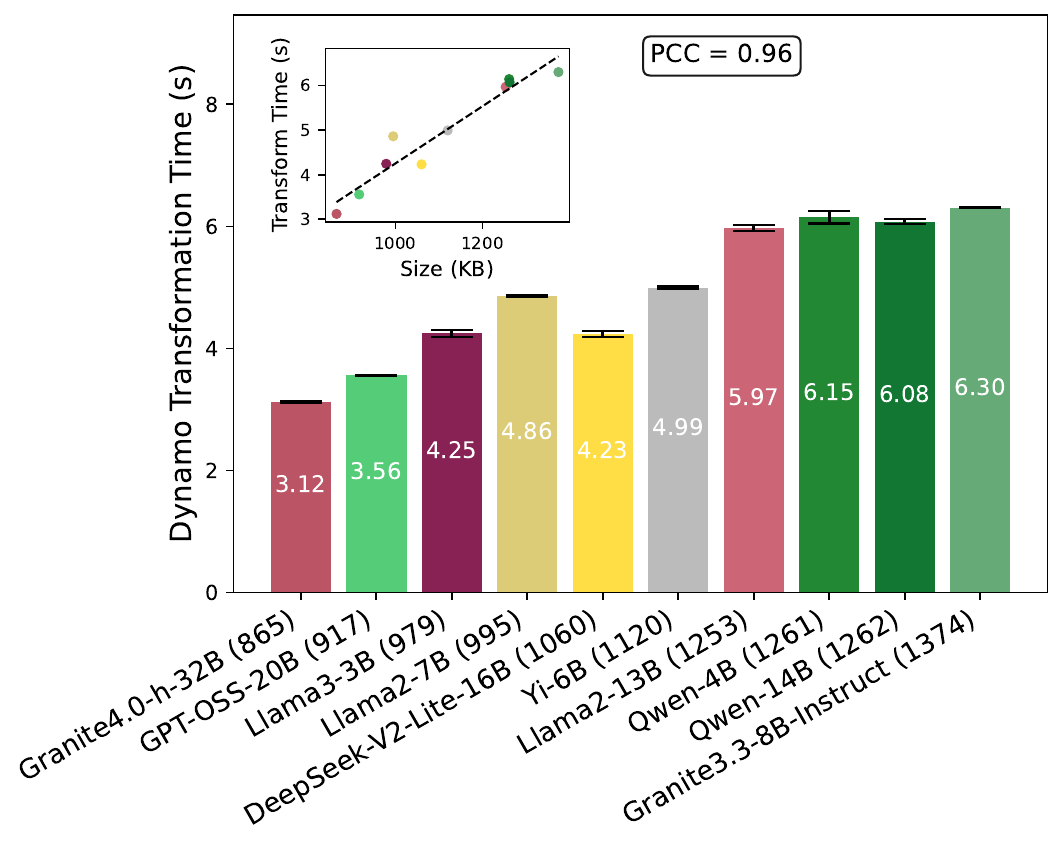}
    \vspace{-0.3cm}
    \caption{Strong linear relationship between compiled graphs size (shown in parentheses, in KB) and Dynamo transformation time.}
    \label{fig:dynamo-transform}
    \vspace{-0.5cm}
\end{figure}

\textbf{Dynamo Bytecode Transformation.}
Dynamo captures and transforms Python bytecode at runtime to extract computational graphs from standard PyTorch programs~\cite{dynamo-overview, dynamo-blog}. In regular execution, each model operation (e.g., matrix multiplication, activation functions, or attention layers) runs as a separate Python call, incurring interpreter overhead and limiting compiler-level optimizations~\cite{dyanmo-paper}. Dynamo instead rewrites the execution into an \emph{intermediate representation (IR)}, a static, compiler-friendly, graph-like form, enabling optimizations such as operator fusion and memory planning~\cite{dyanmo-paper, dynamo-deep-dive, torch-compile-overview}.

We find that Dynamo transformation time grows with the number of layers, as each additional layer introduces more operations that must be traced and transformed. The time also depends on the complexity of each layer, which refers to the number of kernels, metadata and function wrappers that must be traced, meaning that complex layers need more time to be compiled. We observe that a good proxy for the complexity of a layer is the size of the generated compiled graph file. To verify this observation, \cref{fig:dynamo-transform} reports this relationship across models, showing that transformation time increases with the total compiled graph files size, computed as the sum of all layers graph sizes. For example, Llama2-7B and Llama2-13B require 4.86~secs and 5.97~secs respectively. Although they have the same architecture (hence similar complexity), the latter takes more time due to higher number of layers (32 vs. 40).
The inset plot confirms a near-linear scaling trend with PCC = 0.96, indicating that transformation cost is primarily determined by the layer count and per-layer complexity.

\begin{figure}[t]
    \centering
    \includegraphics[width=0.98\linewidth]{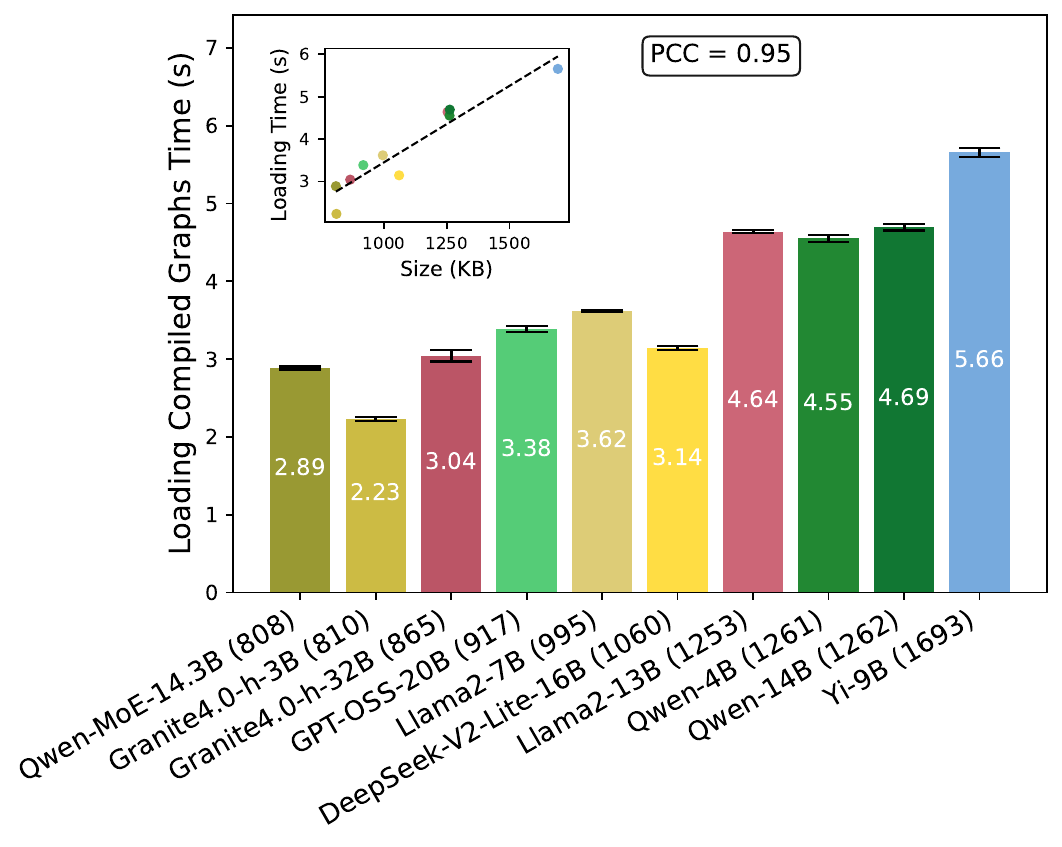}
    \vspace{-0.3cm}
    \caption{Strong linear relationship between compiled graphs size (shown in parentheses, in KB) and loading compiled graphs time.}
    \label{fig:load-graphs}
    \vspace{-0.4cm}
\end{figure}

\paragraph{Loading/Storing Compiled Graphs.}  
After Dynamo generates IR graphs, TorchInductor~\cite{torch-inductor} compiles them into highly optimized, low-level kernels that run efficiently on the GPU~\cite{writing-graph}. vLLM then stores these compiled artifacts in a file system cache, enabling subsequent runs to skip recompilation. Our measurements are performed with this cache already populated; \S\ref{sec:impact-non-cached-graphs} will show the impact of this caching on the startup latency. 

As shown in~\cref{fig:load-graphs}, loading time scales linearly with compiled graph size, from 2.89~secs for Qwen-MoE-14.3B (808~KB) to 5.66~secs for Yi-9B (1.69~MB). Models sharing the same architecture and number of layers, such as Qwen-4B and Qwen-14B, exhibit similar loading times (4.55~secs vs.~4.69~secs) despite different parameter counts. The inset confirms this linear dependence with PCC = 0.95, showing that the loading cost depends primarily on the compiled graph size rather than the model architecture or parameter size.

\summarybox{The \texttt{torch.compile} step time increases with both the number of layers and their complexity, which can effectively be approximated by the total size of the generated compiled graph files.}

\subsection{KVCache Profiling}
\label{sec:kv-cache}

\begin{figure}[t!]
    \centering
    \includegraphics[width=0.98\linewidth]{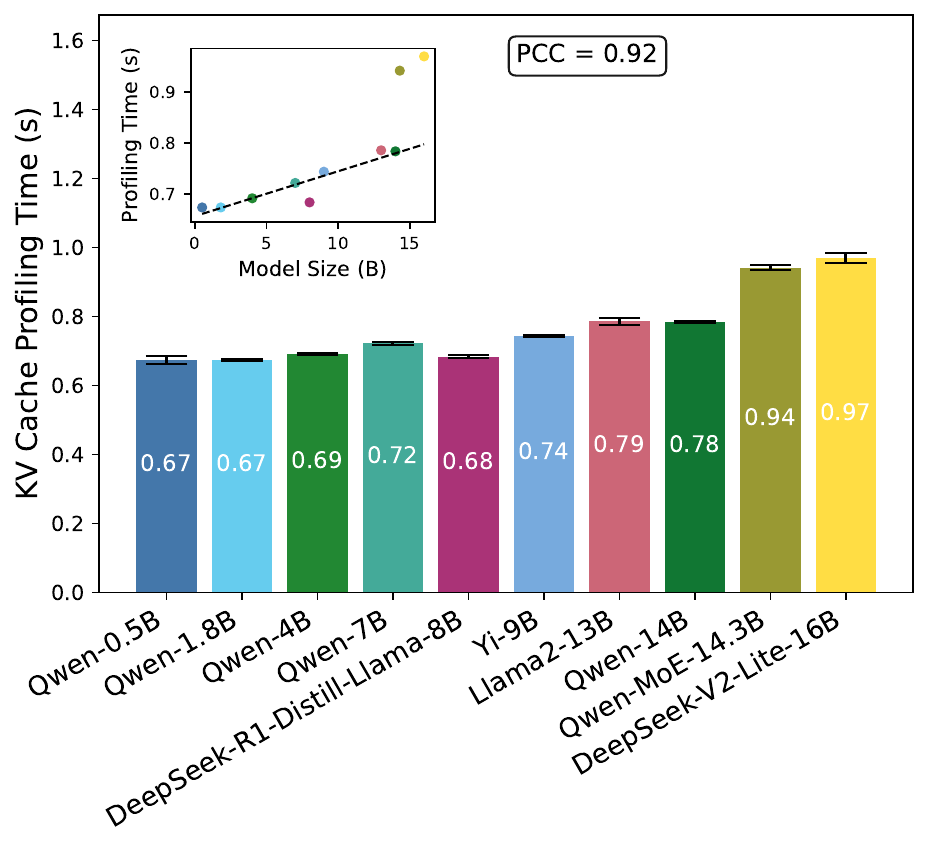}
    \vspace{-0.3cm}
    \caption{KVCache profiling time across different models. The inset shows a linear regression fitted only on non-MoE models (excluding Qwen and DeepSeek, shown as two yellow and olive dots in the top right in the inset) with \texttt{torch.compile} disabled.}
    \label{fig:kv-cache-results}
    \vspace{-0.5cm}
\end{figure}

The fifth step in vLLM’s startup process is key–value (KV) cache profiling, which determines the optimal amount of GPU memory to allocate for the KVCache. The KVCache stores the past key and value tensors generated by the attention mechanism, allowing the model to reuse them efficiently across decoding steps. Since the cache for attention layers grows with each generated token, incorrect allocation can lead to out-of-memory errors. Profiling is therefore critical: vLLM executes a dummy forward pass to measure peak memory usage, and the remaining available GPU memory is then allocated for the KVCache. This ensures a balance between stability and maximum memory utilization during inference.

Since this step involves the first invocation of the model’s forward pass through the dummy run, \texttt{torch.compile} is implicitly triggered, causing the profiling time reported by vLLM to include compilation overhead. Initially, we attempted to isolate the profiling duration by subtracting the measured \texttt{torch.compile} time from the logged values. However, this approach failed to yield a consistent trend. A closer analysis revealed that the inclusion of compilation overhead within the profiling stage distorted the measurements. To obtain a more accurate characterization of the intrinsic profiling behavior, we re-ran this step with \texttt{torch.compile} explicitly disabled, ensuring that only the dummy forward execution time is captured. 

Figure~\ref{fig:kv-cache-results} shows the measured profiling time across models after applying these adjustments. Smaller models such as Qwen-0.5B and Qwen-1.8B require about 0.67~secs, whereas medium-scale models like Qwen-4B and Qwen-7B take 0.69–0.72~secs. Larger transformer models including Yi-9B, Llama2-13B, and Qwen-14B exhibit profiling times between 0.74–0.79~secs, while the heaviest Mixture-of-Experts (MoE) models (Qwen-MoE-14.3B, DeepSeek-V2-Lite-16B) reach up to 0.94–0.97~secs. 
The results confirm that once the compilation overhead is removed, the KVCache profiling time follows a predictable linear trend with model size (PCC = 0.92), except for MoE models. Non-MoE models show a strong linear dependency on model size, consistent with expectations, since this step performs a dummy forward pass whose duration grows proportionally with the number of parameters shown in the inset that excludes fitting for the MoE models. 
In contrast, MoE models deviate due to their dynamic expert activation and load-balancing mechanisms, which introduce additional profiling complexity during the dummy run~\cite{huang2024toward, mu2025comprehensive}. 
Characterizing these non-linearities remains part of our ongoing work.

\summarybox{KVCache profiling scales linearly with the \textit{transformer} model parameter size. MoE models deviate from this trend due to expert routing and varying activation patterns.}

\subsection{CUDA Graph Capturing}
\label{sec:cuda-graph}

\begin{figure}[t]
    \centering
    \includegraphics[width=0.98\linewidth]{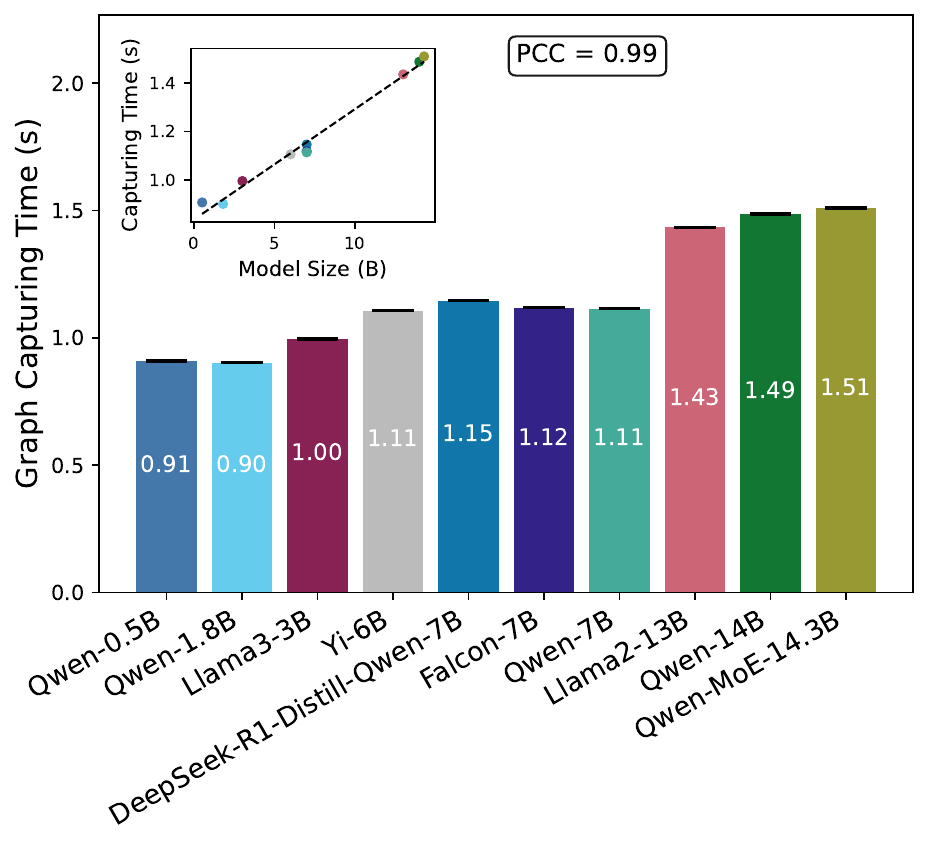}
    \vspace{-0.3cm}
    \caption{Strong linear relationship between model size and CUDA graph capturing time.}
    \label{fig:cuda-graph-size}
    \vspace{-0.3cm}
\end{figure}

The final step of vLLM’s startup process is CUDA graph capturing. During this step, vLLM performs a dummy forward pass to record the execution of inference kernels (including memory allocations, attention computations, and other CUDA operations) into a CUDA graph during the startup process. This graph encodes the exact sequence of GPU operations required for inference. Once captured, the graph can be replayed efficiently without re-launching individual kernels, significantly reducing CPU–GPU synchronization overhead and kernel launch latency. This makes CUDA graph capturing particularly important for achieving high inference throughput and low latency in production environments~\cite{cuda-graph-capture-overview}.

To see the influence of different parameters on this step, we conduct two experiments. \cref{fig:cuda-graph-size} shows the first one, where we measure CUDA capturing time for different models with different sizes. We observe that as the model size increases, so does the capturing time confirming a linear trend with PCC = 0.99. For instance, In~\cref{fig:cuda-graph-size}, Qwen-0.5B takes about 0.91~secs, while Qwen-MoE-14.3B requires around 1.51~secs for capturing. 

\begin{figure}[t]
    \centering
    \includegraphics[width=0.9\linewidth]{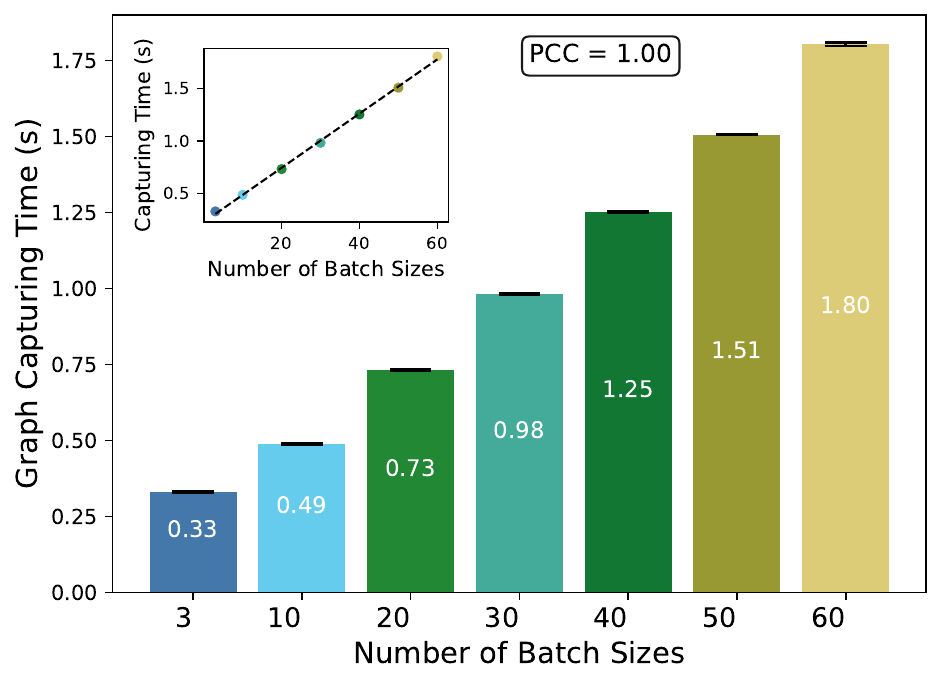}
    \vspace{-0.3cm}
    \caption{CUDA graph capturing time for different batch sizes using the Llama2-7B~\cite{llama2} model.}
    \label{fig:cuda-graph-batch}
    \vspace{-0.5cm}
\end{figure}

In the second experiment, as shown in~\cref{fig:cuda-graph-batch}, we analyze the effect of the batch size on the capturing time. Motivated by the fact that CUDA Graph capturing must be performed separately for each unique batch size~\cite{nvidia-cudagraphs}, we measure the capturing time for a single model, Llama2-7B~\cite{llama2}, using a range of batch sizes.
The results show a familiar trend. The time for capturing the graph increases from 0.33~secs with three batch sizes to 1.8~secs with 60 batch sizes. These numbers reveal a gradual increase in capturing time as both model size and batch size grow, showing a linear relationship with PCC = 1. 

By examining the inset graphs in both figures, we can deduce a clear linear relationship between the capturing time and the two parameters: model size and batch size. In both cases, the graphs demonstrate a steady increase in capturing time as either the model size or batch size increases.

\summarybox{CUDA graph capturing time scales linearly with both model size and the number of batches to be captured.}

\subsection{Summary}
Our breakdown of vLLM’s startup process reveals distinct and quantifiable dependencies across all six foundational steps to vLLM-internal (runtime initialization), and model-dependent (size, complexity, architecture) factors. We believe that these are foundational steps, and new releases of vLLM will not invalidate our insights but will build on them.
Together, these observations demonstrate that each startup step has clear and interpretable scaling trends with respect to specific parameters. This systematic characterization establishes a strong empirical basis for modeling startup latency analytically to help build a predictive scheduler for a serverless autoscaler (\S\ref{sec:analytical-predictor}). 

%% file: src/tables/model-configs.tex
\definecolor{qwen05b}{HTML}{4477AA}
\definecolor{qwen18b}{HTML}{66CCEE}
\definecolor{qwen4b}{HTML}{228833}
\definecolor{qwen7b}{HTML}{44AA99}
\definecolor{qwen14b}{HTML}{117733}
\definecolor{qwen143b}{HTML}{999933}
\definecolor{llama27b}{HTML}{DDCC77}
\definecolor{llama213b}{HTML}{CC6677}
\definecolor{llama33b}{HTML}{882255}
\definecolor{falcon7b}{HTML}{332288}
\definecolor{falcon11b}{HTML}{EE7733}
\definecolor{yi6b}{HTML}{BBBBBB}
\definecolor{yi9b}{HTML}{77AADD}
\definecolor{mistral7b}{HTML}{FFAABB}
\definecolor{gemma7b}{HTML}{99DDFF}
\definecolor{gptoss20b}{HTML}{55CC77}
\definecolor{deepseekv2lite16b}{HTML}{FFDD44}
\definecolor{deepseekr1llama8b}{HTML}{AA3377}
\definecolor{deepseekr1qwen7b}{HTML}{1177AA}
\definecolor{granite33instruct}{HTML}{66AA77}
\definecolor{granite40micro3b}{HTML}{CCBB44}
\definecolor{granite40small32b}{HTML}{BB5566}
\definecolor{opt67b}{HTML}{DDDDDD}
\newcommand{\modelcolor}[1]{\textcolor{#1}{\rule{0.8em}{0.8em}}~}

\begin{table*}[!t]
\centering
\small
\setlength{\tabcolsep}{3pt}
\begin{threeparttable}
\vspace{-0.4cm}
\caption{Comparison of models architectures and configurations. MoE = Mixture of Experts, MHA = Multi-Head Attention, GQA = Grouped Query Attention, MQA = Multi-Query Attention, MLA = Multi-Latent Attention, $K = 1000$. Color boxes correspond to the colors used in all Figures.}
\label{tab:model-configs}
\begin{tabularx}{\textwidth}{p{7.1cm} p{0.9cm} p{0.9cm} p{0.8cm} p{1.1cm} p{1.5cm} p{1cm} p{1.2cm} p{0.9cm}}
\toprule
Model & Layers & Hidden Size & FFN Dim & Heads (Q / KV) & Attention Type & Vocab Size & Tokenizer Size & MoE? \\
\midrule
\modelcolor{llama27b}~LLaMA 2-7B~\cite{llama2} & 32 & 4096 & 11008 & 32 / 32 & Full MHA & 32K & 1.8MB & No \\
\modelcolor{llama213b}~LLaMA 2-13B~\cite{llama2} & 40 & 5120 & 13824 & 40 / 40 & Full MHA & 32K & 1.8MB & No \\
\modelcolor{llama33b}~LLaMA 3-3B~\cite{meta-llama3.2-3b} & 28 & 3072 & 8192 & 24 / 8 & GQA (3:1) & 128K & 8.7MB & No \\
\modelcolor{falcon7b}~Falcon-7B~\cite{falcon-7b} & 32 & 4544 & 18176 & 71 / 1 & MQA & 65K & 2.7MB & No \\
\modelcolor{qwen05b}~Qwen-0.5B~\cite{qwen} & 24 & 1024 & 2816 & 16 / 16 & Full MHA & 152K & 6.8MB & No \\
\modelcolor{qwen18b}~Qwen-1.8B~\cite{qwen} & 24 & 2048 & 5504 & 16 / 16 & Full MHA & 152K & 6.8MB & No \\
\modelcolor{qwen4b}~Qwen-4B~\cite{qwen} & 40 & 2560 & 6912 & 20 / 20 & Full MHA & 152K & 6.8MB & No \\
\modelcolor{qwen7b}~Qwen-7B~\cite{qwen} & 32 & 4096 & 11008 & 32 / 32 & Full MHA & 152K & 6.8MB & No \\
\modelcolor{qwen14b}~Qwen-14B~\cite{qwen} & 40 & 5120 & 13696 & 40 / 40 & Full MHA & 152K & 6.8MB & No \\
\modelcolor{yi6b}~Yi-6B~\cite{ai2024yi} & 32 & 4096 & 11008 & 32 / 4 & GQA (8:1) & 64K & 3.5MB & No \\
\modelcolor{yi9b}~Yi-9B~\cite{ai2024yi} & 48 & 4096 & 11008 & 32 / 4 & GQA (8:1) & 64K & 3.5MB & No \\
\modelcolor{falcon11b}~Falcon-11B~\cite{falcon-11b} & 60 & 4096 & 16384 & 32 / 8 & GQA (4:1) & 65K & 2.7MB & No \\
\modelcolor{mistral7b}~Mistral-7B~\cite{mistral-7b} & 32 & 4096 & 14336 & 32 / 8 & GQA (4:1) & 32K & 1.8MB & No \\
\modelcolor{qwen143b}~Qwen-MoE-14.3B-A2.7B~\cite{qwen-moe} & 24 & 2048 & 1408 & 16 / 16 & Full MHA & 152K & 6.8MB & Yes \\
\modelcolor{gemma7b}~Gemma-7B~\cite{gemma-7b} & 28 & 3072 & 24576 & 16 / 16 & Full MHA & 256K & 17.5MB & No \\
\modelcolor{gptoss20b}~GPT-OSS-20B~\cite{gpt-oss-20b} & 24 & 2880 & 2880 & 64 / 8 & GQA (8:1) & 201K & 27MB & Yes \\
\modelcolor{granite33instruct}~Granite3.3-8B-Instruct~\cite{granite-3.3-8b-instruct} & 40 & 4096 & 12800 & 32 / 8 & GQA (4:1) & 50K & 3.4MB & No \\
\modelcolor{granite40small32b}~Granite4.0-h-small-32B~\cite{granite-4.0-h-small} & 40 & 4096 & 1536 & 32 / 8 & GQA (4:1) & 100K & 6.9MB & Yes \\
\modelcolor{granite40micro3b}~Granite4.0-h-mirco-3B~\cite{granite-4.0-h-micro} & 40 & 2048 & 8192 & 32 / 8 & GQA (4:1) & 100K & 6.9MB & No \\
\modelcolor{deepseekv2lite16b}~DeepSeek-V2-Lite-16B~\cite{deepseekv2} & 27 & 2048 & 1408 & 16 / 16 & MLA & 102K & 4.4MB & Yes \\
\modelcolor{deepseekr1llama8b}~DeepSeek-R1-Distill-Llama-8B~\cite{deepseekr1} & 32 & 4096 & 14336 & 32 / 8 & GQA (4:1) & 128K & 8.7MB & No \\
\modelcolor{deepseekr1qwen7b}~DeepSeek-R1-Distill-Qwen-7B~\cite{deepseekr1} & 28 & 3584 & 18944 & 28 / 4 & GQA (7:1) & 152K & 6.8MB & No \\
\bottomrule
\end{tabularx}
\end{threeparttable}
\end{table*}

%% file: src/tables/system-env.tex
\begin{table*}[!ht]
\small
\setlength{\tabcolsep}{3pt}
\centering
\vspace{-0.4cm}
\caption{Hardware and software configurations for the nodes used in experiments.}
\label{tab:system-env}
\begin{tabular}{lllll}
\toprule
 & node1 (\texttt{n1)} & node2 (\texttt{n2}) & node3 (\texttt{n3}) & node4 (\texttt{n4}) \\
\midrule
CPU & AMD EPYC 9354 (32C) & AMD EPYC 9354 (32C) & 2x Intel Xeon Platinum 8568Y+ (2x48C)  & 2x Intel Xeon Gold 5520+ (2x28C) \\
GPU & H100 NVL & L40S & H100 & L40S \\
DRAM & DDR5 251GB & DDR5 251GB & DDR5 2TB & DDR5 2TB \\
OS & Debian 12 & Debian 12 & Red Hat Enterprise Linux (RHEL) 9 & Red Hat Enterprise Linux (RHEL) 9 \\
Kernel & \texttt{6.1.0-40-amd64} & \texttt{6.1.0-40-amd64} & \texttt{5.14.0-503.34.1.el9\_5} & \texttt{5.14.0-503.34.1.el9\_5}\\
Python & 3.11.2 &3.11.2  & 3.12 & 3.12 \\
CUDA & 12.6 (Driver 580.82.07) & 12.6 (Driver 580.82.07) & 12.8 (Driver 570.124.06) & 12.8 (Driver 570.124.06) \\
PyTorch & 2.7.1+cu126 & 2.7.1+cu126 & 2.7.1+cu126 & 2.7.1+cu126\\
vLLM & v0.10.1.1 & v0.10.1.1 & v0.10.1.1 &  v0.10.1.1 \\
SSD & -- & -- & 4x PCIe 5.0 SSDs & -- \\ 
FS & -- & -- & XFS/LVM with RAID-0 mirror & -- \\ 
\bottomrule
\vspace{-0.5cm}
\end{tabular}
\end{table*}

%% file: src/impact.tex
\section{Impact of Benchmarking Environment}
\label{sec:env-config}
In this section, we analyze the impact of the benchmarking environment (e.g., GPU, CPU, storage) and different configurations on our findings from \S\ref{sec:experiments}. We also examined additional factors----including containerization with Docker, varying PyTorch and Python versions, and modifying common runtime configuration flags (e.g., \texttt{--max-model-len} and OpenAI-compatible vLLM arguments)----but observed no measurable or statistically significant impact on startup time and are thus omitted from discussion here.

\subsection{Impact of Different GPUs}
\label{sec:impact-gpu}

\begin{figure}[t]
    \centering
    \includegraphics[width=0.98\linewidth]{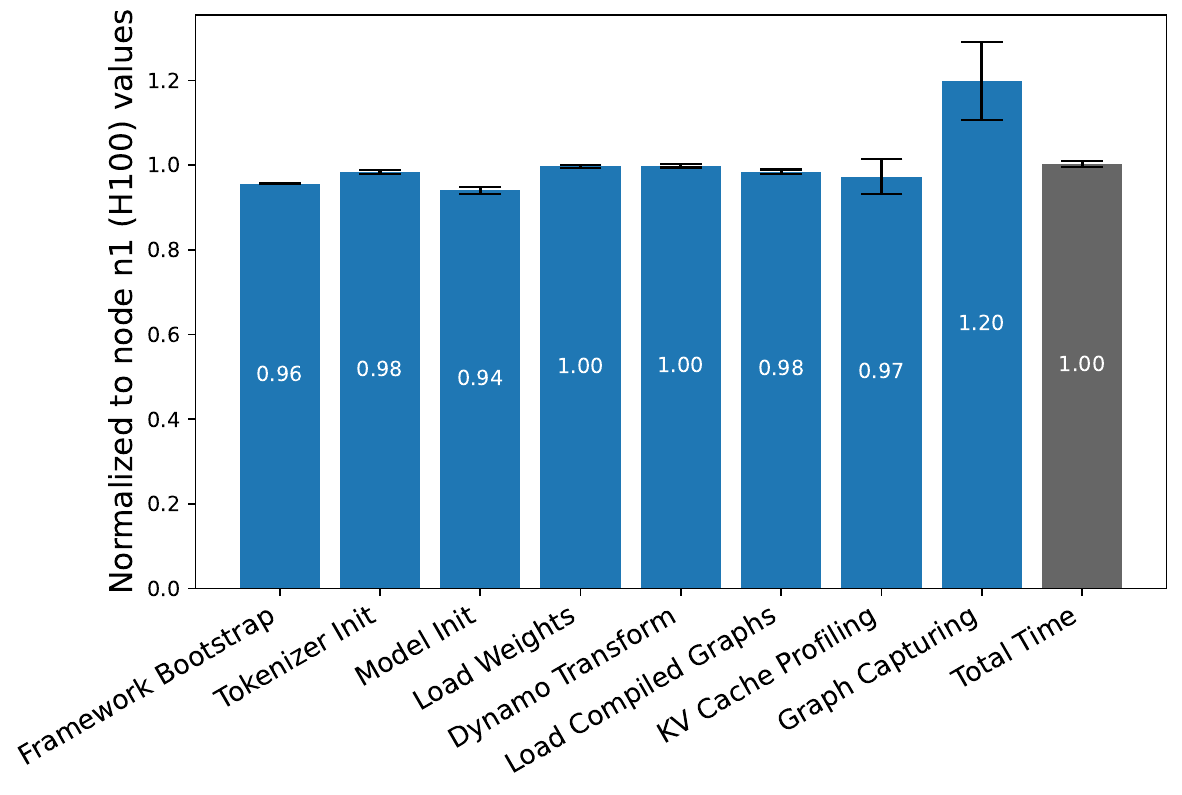}
    \vspace{-0.3cm}
    \caption{Startup steps comparison between H100 (n1) and L40S (n2) GPUs. All values are normalized to the H100 baseline.}
    \label{fig:h100-vs-l40s}
\end{figure}

To validate the resource dependency findings presented in~\cref{fig:vllm-init}, we repeat the previous experiments on node \texttt{n2}, which has the same system configuration as \texttt{n1} but with L40S GPU instead of H100 (see~\cref{tab:system-env}). These experiments are done on the first ten models listed in~\cref{tab:model-configs}. \cref{fig:h100-vs-l40s} illustrates the average speedup of each startup step across all evaluated models, calculated as the ratio of startup time on the H100 to the L40S (the $y$-axis).

As shown in the figure, most steps exhibit almost no speedup, indicating negligible performance gains when using the H100 over the L40S. The only notable exception is CUDA Graph Capturing, which demonstrates a speedup of 1.2$\times$ due to the forward pass run on the GPU during this step. These results are consistent with our earlier observations in~\cref{fig:vllm-init}, reinforcing the conclusion that the startup process is predominantly CPU-bound. Notably, despite the different GPU architecture, and the H100’s significantly higher theoretical TFLOPs throughput compared to the L40S,~\cite{h100-vs-l40s-1, h100-vs-l40s-2}, this advantage does not translate into faster overall startup time. This observation indicates that the GPU performance has limited impact on the startup process.

\begin{figure}[t]
\centering
\includegraphics[width=0.98\linewidth]{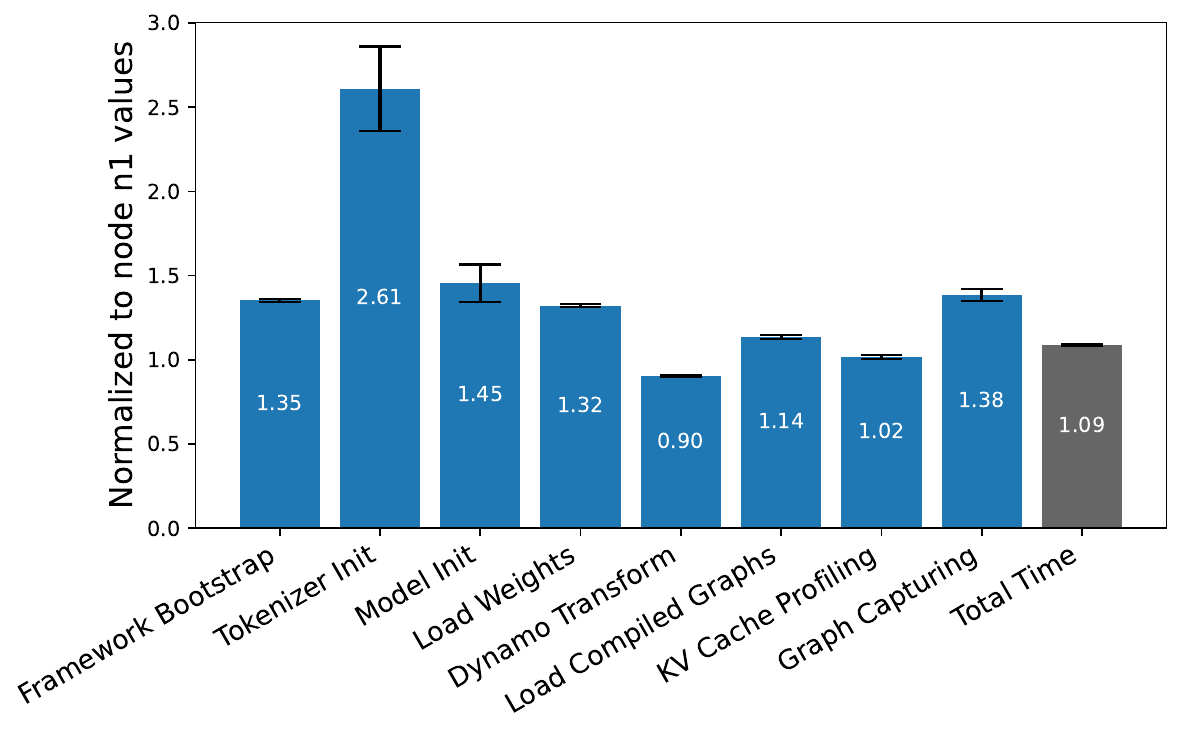}
\vspace{-0.3cm}
\caption{Comparison between AMD~EPYC~9354 (\texttt{n1}) and Intel~Xeon~Platinum~8568Y+ (\texttt{n3}) CPUs with H100 GPUs.}
\vspace{-0.5cm}
\label{fig:cpu_h100}
\end{figure}

\subsection{Impact of Different CPUs}
\label{sec:impact-cpu}
To further validate that vLLM's startup process is predominantly CPU-bound, we conduct experiments comparing environments with identical GPUs but different CPUs.
Specifically, we compare our main node \texttt{n1} (AMD EPYC~9354 + NVIDIA H100), against node \texttt{n3}, equipped with an Intel Xeon Platinum~8568Y+ CPU and H100 GPU.

As shown in~\cref{fig:cpu_h100}, changing the CPU has a noticeably higher impact on startup time than changing the GPU (\S\ref{sec:impact-gpu}), reinforcing our earlier conclusion that the startup process is largely CPU-bound. However, the direction and magnitude of speedups across substeps varied considerably. For example, \texttt{n1} outperforms \texttt{n3} in \textit{Tokenizer Initialization}, and \textit{Model Initialization}, while \texttt{n3} is faster in \textit{Graph Capturing} and \textit{Dynamo Transformation}. The same experiment is also conducted between node \texttt{n2} and another node \texttt{n4}, equipped with Intel Xeon Gold~5520+ CPU and the same L40S GPU. Similarly, relative performance across steps fluctuates rather than following a consistent trend; the corresponding figure is omitted for space. 

\begin{figure}[t]
    \centering
    \includegraphics[width=0.9\linewidth]{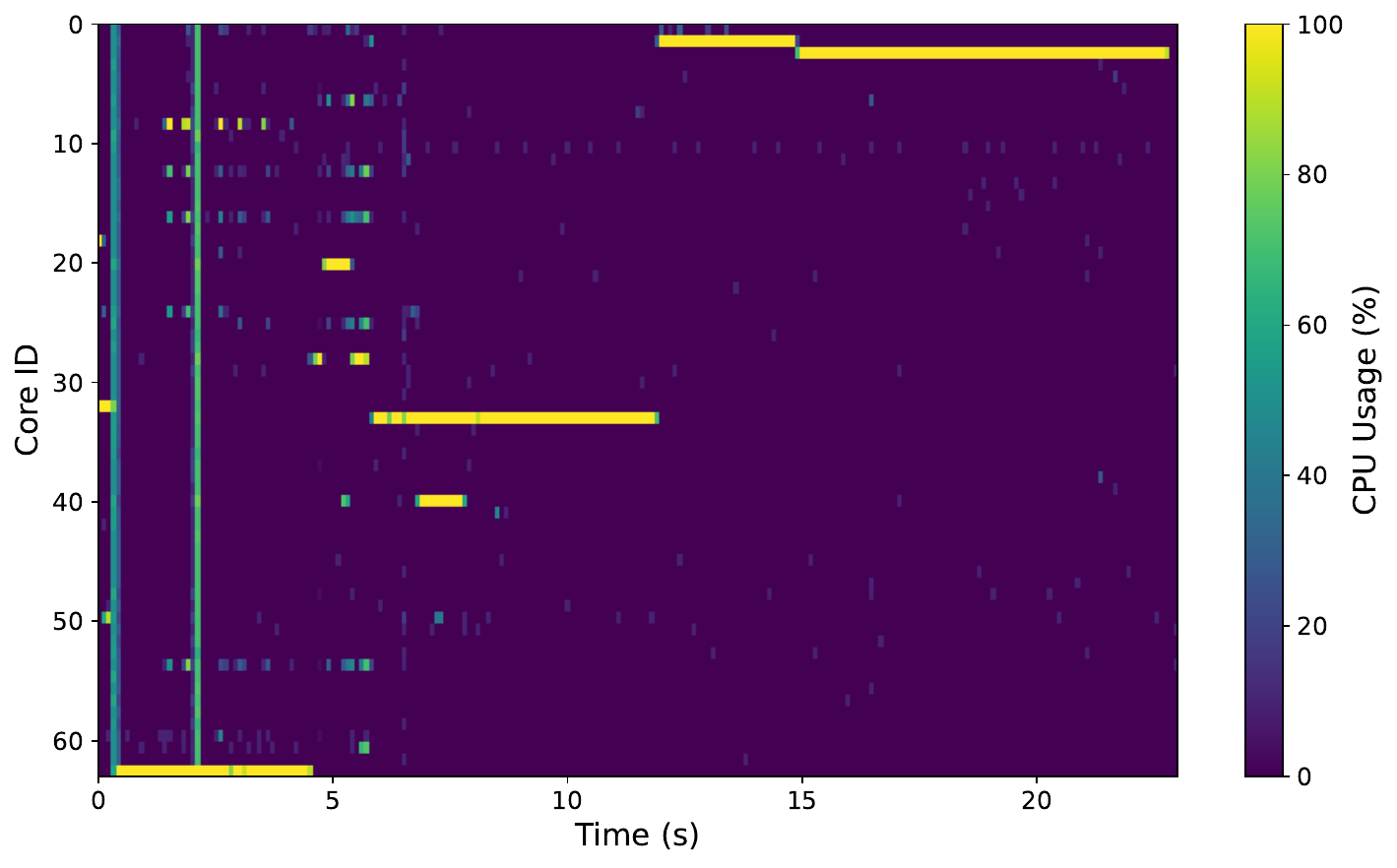}
    \caption{CPU usage per core over time during vLLM startup for the Qwen-4B model. Sampling interval: 100\,ms.}
    \label{fig:cpu-usage-qwen4b}
    \vspace{-0.5cm}
\end{figure}

To better understand CPU involvement during the startup process, we monitored the utilization of individual CPU cores over time using a sampling interval of 100\,ms. \cref{fig:cpu-usage-qwen4b} shows the per-core utilization during the startup of the Qwen-4B~\cite{qwen} model.
The heatmap shows that at any given time, at least one CPU core reaches full (100\%) utilization, indicating that vLLM continuously keeps one core saturated throughout the startup process. In contrast, there are very few instances where more than two cores are simultaneously fully utilized, suggesting that most startup operations are sequential or involve limited parallelism.

Overall, while these results confirm that CPU choice has a substantial effect on the vLLM startup latency, a precise attribution of the performance differences to the CPU (micro)architectural and systems-level factors (scheduling, runtime) remains part of our ongoing analysis.

\subsection{Impact of SSDs}
\label{sec:impact-ssd}

\begin{figure}[t]
    \centering
    \includegraphics[width=0.98\linewidth]{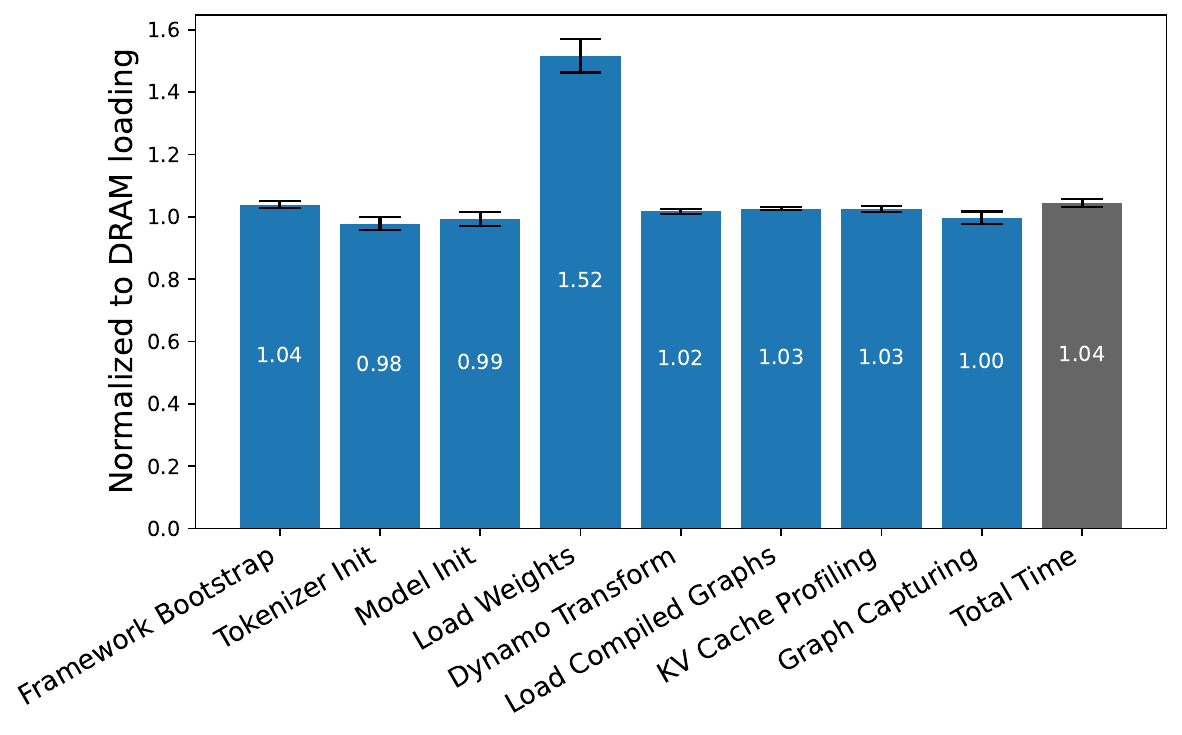}
    \vspace{-0.3cm}
    \caption{Impact of running the startup process while the model weights are retrieved from storage (SSD).}
    \label{fig:ssd}
\end{figure}

Our previous experiments used model weights and other configuration files cached in the warm Linux buffer memory (DRAM). To study the impact of loading directly from storage (SSDs), we repeat the experiments after flushing the buffer cache between runs, forcing all reads to be from the SSD. We conduct these experiments on node \texttt{n3}, equipped with an LVM volume spanning with mirroring four PCIe 5.0 SSDs, delivering read and write throughput of 25 and 15 GB/s, respectively with \texttt{fio} for large I/O transfers. In this experiment, we use the first ten models listed in~\cref{tab:model-configs}, each repeated five times to account for variations. \cref{fig:ssd} summarizes the averaged results across models, normalized to the DRAM baseline.

As expected, the only step that has a significant impact due to the data loading from SSDs is the Model Loading step. It slows down by a factor of 0.5$\times$. The use of SSDs has minimal impact on all other steps, indicating that storage I/O does not significantly affect CPU-bound steps such as compilation or profiling. However, despite this considerable relative improvement in the weight-loading step, the overall startup time improves by only 1.04$\times$. This modest total gain arises because the loading step constitutes only about 7–10\% of the total startup duration in our measurements.

\begin{figure}[t]
    \centering
    \includegraphics[width=0.95\linewidth]{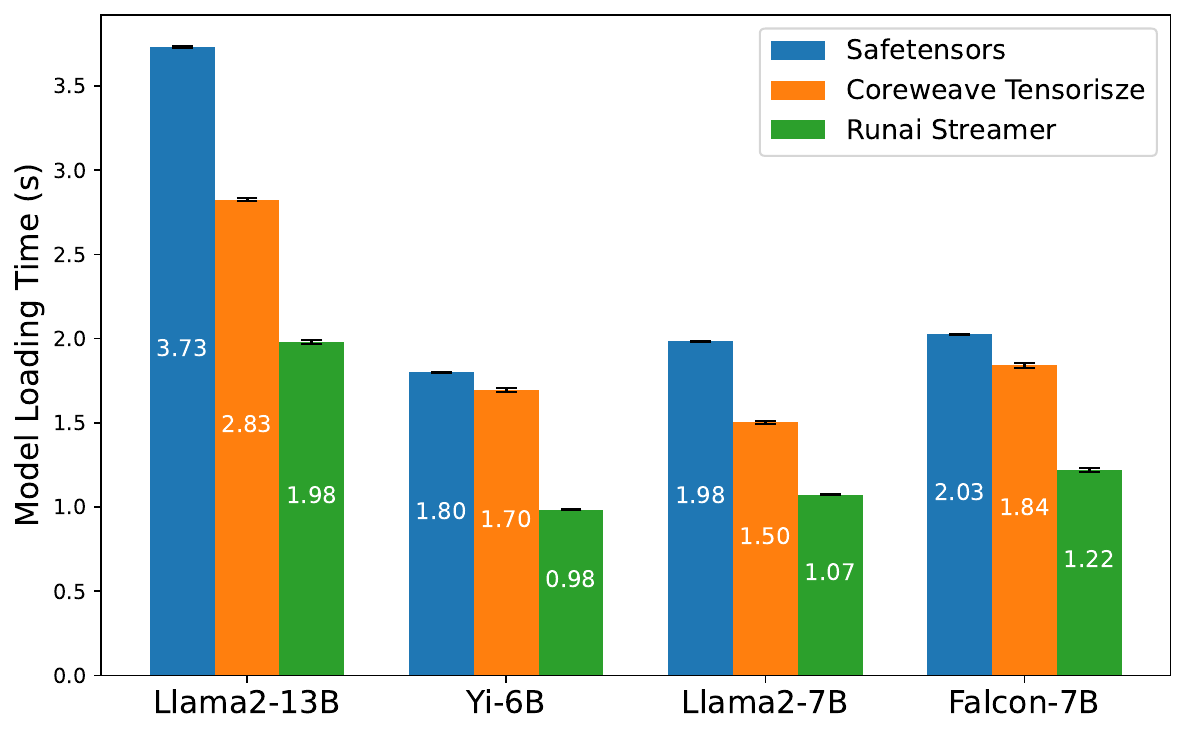}
    \vspace{-0.3cm}
    \caption{Model loading times across four models using different loading backends.}
    \vspace{-0.5cm}
    \label{fig:diff-loading-methods}
\end{figure}

\subsection{Impact of Model Weights' Loading Methods}
vLLM supports multiple methods for loading model weights, each adopting distinct serialization and deserialization strategies. To understand how these methods affect startup latency, we evaluate three supported baselines: 
(i) \textbf{Safetensors} is the default method used in our previous experiments, and it serves as the baseline~\cite{safetensors}. It stores model tensors as pre-serialized binary files that are memory-mapped and loaded directly into CPU memory before GPU transfers.
(ii) \textbf{Run:ai Model Streamer} enables concurrent reading and streaming of tensors directly into GPU memory, reducing I/O bottlenecks~\cite{runai}.
(iii) \textbf{CoreWeave Tensorizer} uses a custom serialization format optimized for fast deserialization, allowing tensors to load directly into GPU without intermediate CPU staging~\cite{tensorize}.

As shown in~\cref{fig:diff-loading-methods}, the \textit{Model Loading} step exhibits substantial variation across different loading methods. Tensorizer consistently achieves the lowest latency, loading models up to 53–60\% of Safetensors' time, while Run:ai Model Streamer provides moderate gains due to overlapping I/O and GPU transfers. These results demonstrate that weight loading is one of the few I/O-sensitive components of the startup process, and that optimizing data streaming and deserialization can yield meaningful reductions in the total startup time.

\begin{figure}[t]
    \centering
    \includegraphics[width=0.98\linewidth]{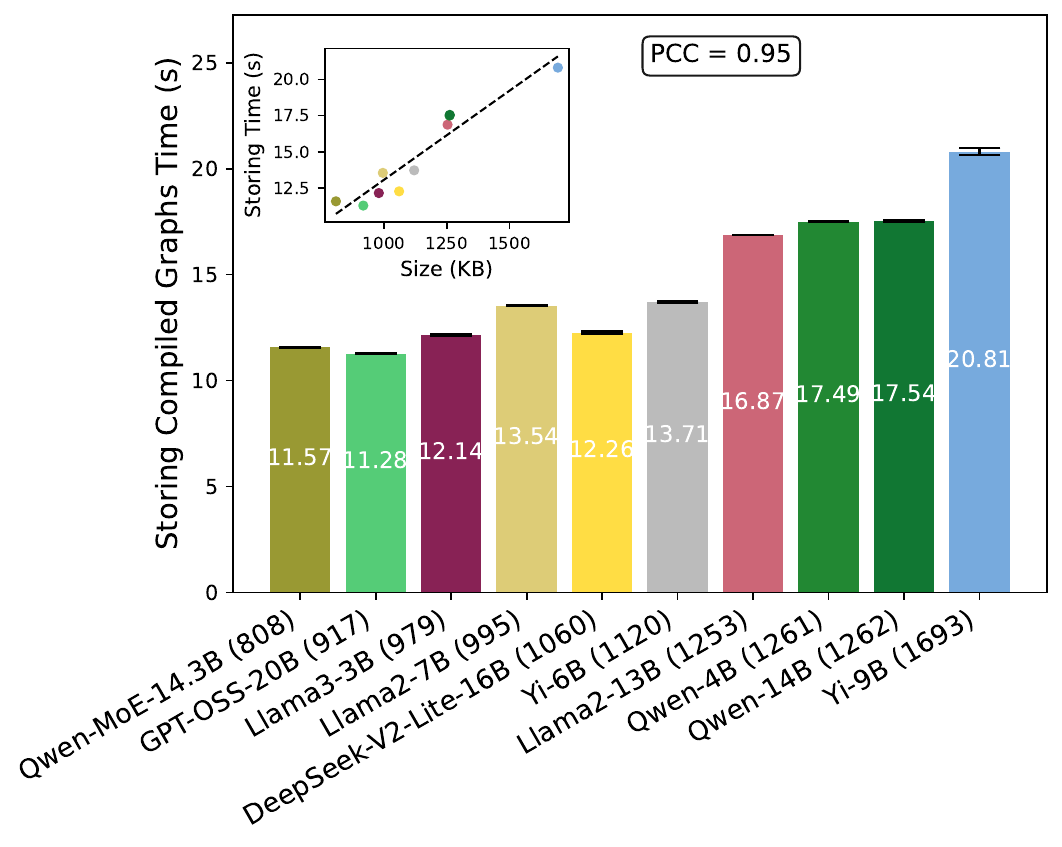}
    \caption{Strong linear relationship between compiled graphs size
(shown in parentheses) and storing compiled graphs time. }
    \label{fig:store-graphs-results}
\end{figure}

\subsection{Impact of Non-Cached Compiled Graphs}
\label{sec:impact-non-cached-graphs}
By default, vLLM caches the compiled computation graphs generated during the \texttt{torch.compile} step after the first run, enabling subsequent runs to bypass costly graph transformations. To quantify the cost of a completely cold start, we disable this cache by setting the environment variable \texttt{VLLM\_DISABLE\_COMPILE\_CACHE=1}, forcing vLLM to regenerate and store all compiled graphs at runtime. As shown in~\cref{fig:store-graphs-results}, disabling the cache dramatically increases the latency of this step. The total graph storing time ranges between 11–21~secs across models, compared to 3–6~secs when cached (see~\cref{fig:load-graphs}). Furthermore, as observed earlier, the compilation cost scales nearly linearly (PCC = 0.95) with the size of the compiled graphs, similar to our results in \S\ref{sec:torch-compilation}.

%% file: src/modeling2.tex
\section{Analytical Predictor}
\label{sec:analytical-predictor}
\begin{figure}[t]
    \centering
    \includegraphics[width=0.98\linewidth]{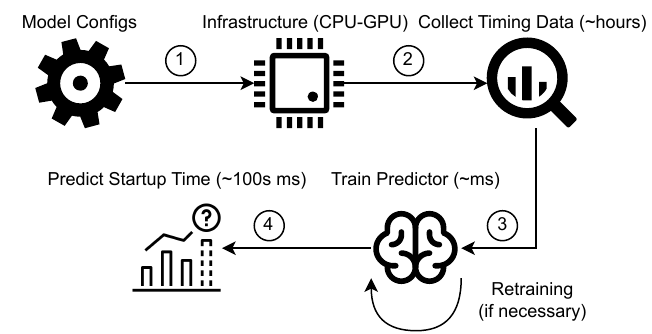}
    \caption{Workflow of the proposed predictor.}
    \label{fig:predictor-overview}
    \vspace{-0.5cm}
\end{figure}

Building on our detailed breakdown of vLLM’s startup process, we now introduce a white-box regression-based analytical predictor for non-MoE models. This predictor estimates startup latency based on the model configurations and environment characteristics that we studied in \S\ref{sec:experiments} and \S\ref{sec:env-config}. \Cref{fig:predictor-overview} shows the overall working of the predictor consisting of the following four steps: (i) gathering of model configuration information for desired models; (ii) automated running of vLLM with models to collect the startup latency timing data from logs (takes typically a few hours); (iii) training step-specific predictors for the environment; and (iv) predicting the startup time.

Beyond estimating cold-start latency, the predictor serves as an interpretable conceptual model of the startup process that practitioners can use to reason about vLLM's startup behavior. For example, when diagnosing performance regressions or unexpected results, such as~\cite{discussion_2}, the predictor provides a step-wise baseline against which empirical measurements can be compared, helping to identify the stages responsible for the slowdown. In addition, its step-wise structure enables estimating the performance of optimizations that reuse parts of the initialization path. For instance, we confirm that the predictor can be used to estimate the performance of fast model re-initialization mechanisms such as vLLM’s sleep mode, and that its estimates match the performance trends reported by the vLLM developers~\cite{vllm-sleep}.

\textbf{Design Rationale.} Our key insight from \S\ref{sec:experiments} is that each startup step exhibits a simple and often near-linear dependency on its corresponding parameters, such as model size for weight loading and graph size for compilation. Leveraging this observation, we design a \emph{white-box decomposed predictor}: a lightweight regressor for each startup step, trained independently using linear regression. This modular formulation preserves interpretability, as the contribution of each parameter remains explicitly visible in its respective step model, while collectively achieving strong predictive accuracy when aggregating across steps. 

An alternative approach here would be to build a monolithic black-box predictor that jointly captures the dependencies between all model- and vLLM-specific parameters. While such a global neural predictor could, in principle, learn these intricate interactions, it would increase complexity, and lack interpretability, thus making it difficult to attribute latency variations to specific model factors or startup steps~\cite{molnar2020interpretable, lipton2018mythos}. In contrast, our white-box approach enables us to retrain only specific step-specific regressors when a new model or hardware becomes available. 

\begin{figure}[t]
    \centering
    \includegraphics[width=0.85\linewidth]{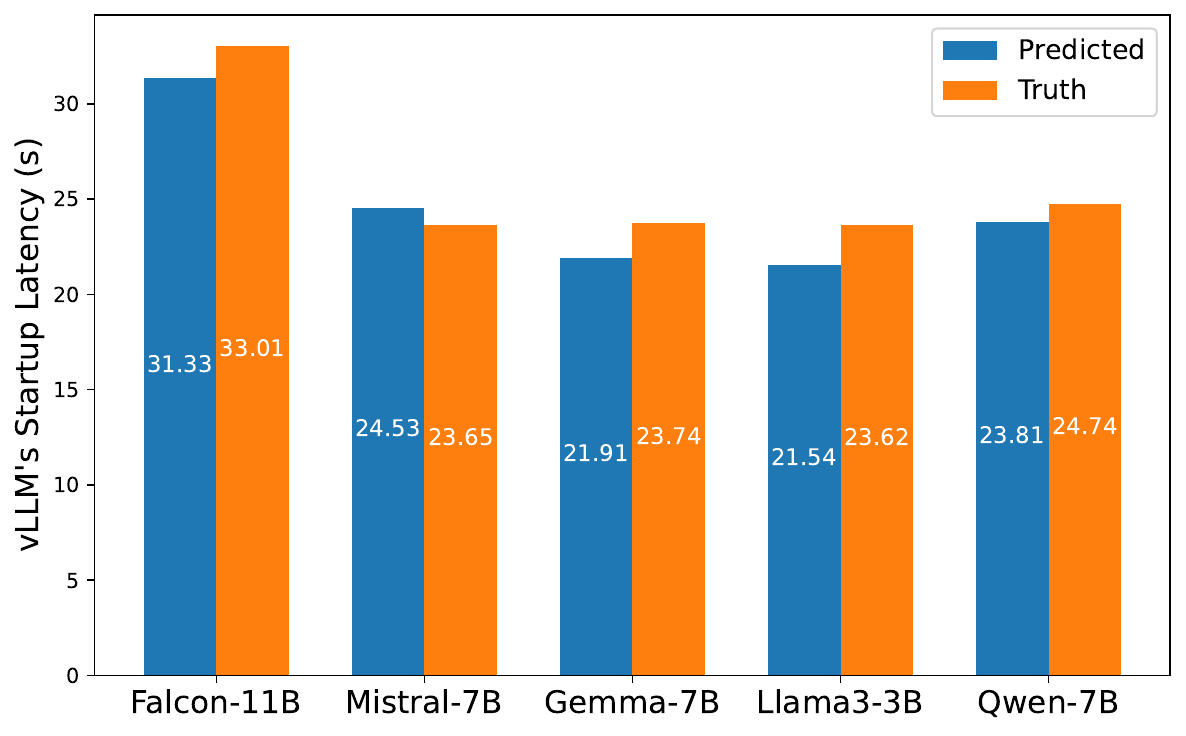}
    \vspace{-0.3cm}
    \caption{Validation of the predictor against measured startup latency across different models.}
    \label{fig:predictor-validation}
    \vspace{-0.5cm}
\end{figure}

\textbf{Validation.} We report the accuracy of the predictor. We use all non-MoE models listed in~\cref{tab:model-configs}, reserving Falcon-11B, Gemma-7B, Mistral-7B, Llama3-3B and Qwen-7B for validation and the remaining models for training the predictors. For each model, we ran the vLLM startup process five times, averaged the results, and used the data to fit the step-specific predictors. The final predictions are compared against the measured values in~\cref{fig:predictor-validation}. Despite its simplicity, the predictor achieves strong accuracy, with a mean squared error (MSE) of~2.42~secs and a maximum error of~2.08~secs, observed for Llama3-3B.
We have also run and validate the predictor on \texttt{v0.11}, released on Oct 2$^{nd}$, and report that our approach and predictor are still accurate (MSE of~2.62~secs), further validating the approach taken in this work for the analysis and modeling of vLLM's startup process. 

%% file: src/new-discussion.tex
\section{Discussion}
 \label{sec:discussion}

\textbf{Benchmarking Scope.}
In a distributed inference service (e.g., serverless inference), cold-start latency is the result of multiple interacting factors:
(i) distributed-environment overheads, such as networking, remote storage accesses, and container retrieval;
(ii) node-local dynamics, including PCIe contention, OS scheduling, and container initialization; and
(iii) the inference engine startup itself.
Our study focuses on factor (iii), the vLLM engine initialization. We design our experimental setups so that the engine startup becomes the dominant bottleneck, thereby minimizing noise from external system components. The hardware and software stack used in our evaluation reflects a typical mid- to high-end GenAI deployment, where CPU-side processing is the primary contributor to startup latency.
While factors (i) and (ii) are important in real-world distributed environments, they are orthogonal to the engine-level dynamics analyzed in this work. A full end-to-end characterization that integrates all factors is an important direction for future research.

\textbf{Longevity of Insights.}
 vLLM evolves rapidly, with frequent architectural changes that may alter the relative cost of individual startup steps. While such changes may require retraining of affected step-specific regressors, the modular nature of our predictor confines this retraining to localized components rather than the entire model. More importantly, our primary contribution lies in the methodology used to decompose and analyze the startup process, rather than in any specific parameter values. The six foundational steps identified in~\cref{fig:vllm-init} capture fundamental operations that are inherent to modern inference engines. Even as vLLM continues to evolve, these stages and their relationships to underlying computational and I/O factors remain grounded in persistent system principles. As a result, we expect the decomposition and analysis methodology to remain applicable, with only parameter re-tuning required to adapt to future vLLM versions.
 
\textbf{Generalizability.}
The identified six steps correspond to the core initialization stages that are present across post-V1 vLLM releases, and similar startup phases have been reported in concurrent analyses of containerized inference services~\cite{llm-d-benchmark}. Moreover, our experimental findings remain consistent across a diverse set of configurations, including two GPU types, two CPU platforms, 22 models, and multiple versions of Python and PyTorch. These results suggest that the observed scaling trends and step-level behaviors are not tied to a single hardware or software stack, but instead reflect more general properties of modern LLM inference systems.
 
\textbf{Assumption of Linearity.}
Our predictors assume predominantly linear relationships between model parameters and step latency. While justified by empirical evidences in \S\ref{sec:experiments}, certain nonlinear behaviors found in MoE, SSM-transformer hybrid, diffusion models require more expressive analysis to capture their performance profile. Fortunately, this analysis is restricted to the KVCache Profiling step only.

\textbf{Measurement Noise.}
 The accuracy of the proposed predictor is inherently tied to the characteristics of the underlying CPU–GPU hardware and the conditions under which the data was collected. Predictors may need to be re-trained when deployed on substantially different infrastructures. Furthermore, training data was gathered under controlled experimental conditions; real-world deployments may experience background workload interference, NUMA effects, or scheduler contention that introduce additional noise not captured by our measurements. Additionally, timing analysis relies on vLLM’s internal logging granularity, which can add up to hundreds of milliseconds of measurement error. To mitigate this, we averaged results over multiple runs, yet minor deviations may remain.
 

%% file: src/related-work.tex
\section{Related Work}\label{sec:rwork}

Prior studies have explored cold start latency in the context of both traditional serverless platforms and emerging LLM inference systems. Early works such as SAND~\cite{akkus2018sand} and SEUSS~\cite{fuerst2021faascache} addressed general-purpose function startup overhead through lightweight containerization and runtime reuse. Other works in this domain focus on forecasting or avoiding cold start \emph{occurrences}, rather than predicting their \emph{duration}~\cite{shiekhani2025hybrid, golec2024cold,jegannathan2022time,nguyen2025taming}.
More recent efforts around LLMs, including Sarathi-Serve~\cite{agrawal2024taming}, Llumnix~\cite{sun2024llumnix}, and DistServe~\cite{zhong2024distserve}, investigate the throughput-latency tradeoff in large-scale model serving but primarily focus on steady-state inference rather than startup costs.
With the emergence of scalable containerized LLM services, several works target cold start latency as a first‐class optimization goal. For example, ParaServe~\cite{paraserve} focuses on reducing model fetching and initialization delays in serverless LLM serving by overlapping parameter loading across GPU servers and exploiting pipeline parallelism. TIDAL uses fine‐grained execution tracing to generate adaptive templates that bypass much of the cold start overhead in serverless settings~\cite{cui2025efficient}. ServerlessLLM introduces techniques such as multi‐tier checkpoint loading, locality‐aware scheduling, and live migration to minimize delays before readiness~\cite{fu2024serverlessllm}. CSGO targets cold start latency in edge‐distributed LLM systems by dynamically partitioning models and overlapping model loading, computation, and communication~\cite{liu2025csgo}. Medusa accelerates serverless LLM inference by materializing frequently used execution states, enabling rapid function startup and reuse across invocations~\cite{zeng2025medusa}.

In comparison to the past literature, our work is the first to systematically decompose and analyze the startup process of vLLM, a popular open-source inference engine used widely. While prior works examined isolated factors such as model loading or CUDA graph capturing, we provide a holistic breakdown across all startup steps and demonstrate the predominance of CPU-bound bottlenecks. Furthermore, our analytical predictor extends prior measurement studies by offering a practical, interpretable model for predicting cold start latency under varying hardware and software configurations. To the best of our knowledge,
this is the first predictor that estimates the startup latency of
an LLM inference engine in an interpretable and modular
manner.

%% file: src/conclusion.tex
\section{Conclusion}
\label{sec:conclusion}

In this paper, we presented the first systematic characterization of \texttt{vLLM}'s startup latency. By decomposing the startup process into six key steps, we identified dominant bottlenecks and quantified their dependence on different model and hardware configurations. Our analysis showed that startup latency is largely CPU bounded and only marginally affected by GPU performance.
Building on these findings, we proposed a regressor-based modular analytical predictor that estimates startup latency with high accuracy using model and system parameters.

In future work, we plan to integrate this predictor into real-world serverless orchestration frameworks to enable proactive cold start mitigation in multi-tenant environments. Additionally, our step-wise decomposition opens directions for further optimization, such as overlapping or parallelizing steps like \texttt{torch.compile}, KVCache profiling, and CUDA graph capturing. These extensions can help reduce startup latency even further, advancing toward adaptive, scalable, and low-latency LLM serving systems.

%% file: src/appendix.tex
\section{Artifact Appendix}

\subsection{Abstract}

The artifact contains code and scripts required to reproduce the profiling experiments presented in the paper. The artifact is based on vLLM \texttt{v0.10.1.1} with additional profiling logs. It includes scripts to download all evaluated LLMs mentioned in \cref{tab:model-configs}, apply modifications to the vLLM runtime to include additional logs, and reproduce all figures from the paper automatically.

The provided workflow installs dependencies, downloads the required models, and executes figure-specific scripts to reproduce the experimental results. Each figure can be reproduced using a unified script interface. Some figures require multiple GPUs or machines to match the experimental setup.

\subsection{Artifact check-list (meta-information)}

{\small
\begin{itemize}
\item {\textbf{Program:} Python, vLLM-based profiling scripts}
  \item {\textbf{Binary:} Pretrained LLM models}
  \item {\textbf{Model:} Multiple HuggingFace LLMs ($\sim$600GB total)}
  \item {\textbf{Run-time environment:} Linux, Python 3.11, CUDA 12.x}
  \item {\textbf{Hardware:} NVIDIA H100, L40S GPUs; AMD EPYC and Intel Xeon CPUs}
  \item {\textbf{Execution:} Bash \& Python scripts per figure}
  \item {\textbf{Metrics:} Throughput, latency, memory, and loading times}
  \item {\textbf{Output:} PDF figures}
  \item {\textbf{Experiments:} Figure reproduction scripts}
  \item {\textbf{How much disk space required (approximately)?:} $\sim$600GB}
  \item {\textbf{How much time is needed to prepare workflow (approximately)?:} 1–2 hours (model downloads)}
  \item {\textbf{How much time is needed to complete experiments (approximately)?:} Several hours to a day depending on hardware}
  \item {\textbf{Publicly available?:} Yes}
  \item {\textbf{Code licenses (if publicly available)?:} Apache 2.0}
  \item {\textbf{Data licenses (if publicly available)?:} HuggingFace model licenses}
  \item {\textbf{Workflow automation framework used?:} Bash and Python scripts}
  \item {\textbf{Archived (provide DOI)?:} 10.5281/zenodo.19591523}
\end{itemize}
}

\subsection{Description}

\subsubsection{How to access}

The artifact is publicly available on GitHub:

\href{https://github.com/upb-cn/vllm-startup-profiler}{https://github.com/upb-cn/vllm-startup-profiler}

\subsubsection{Hardware \& Software dependencies}
The experiments were conducted on multiple GPUs and CPUs. The detailed configurations used are summarized in \cref{tab:system-env}

\subsubsection{Models}

The artifact uses pretrained LLM weights downloaded from HuggingFace. The full set of models requires approximately 600\,GB of disk space. A HuggingFace access token is required, and some models may require manual access approval.

Models are downloaded via:
\begin{verbatim}
python3 download_models.py <hf_token>
\end{verbatim}

\subsection{Installation}

\begin{enumerate}
    \item Clone the repository:

\verb|git clone| \url{https://github.com/upb-cn/vllm-startup-profiler}
\verb|cd vllm-startup-profiler|

    \item Install dependencies:
\begin{verbatim}
pip install -r requirements.txt
\end{verbatim}

    \item Apply custom vLLM modifications:
\begin{verbatim}
python3 apply_vllm_changes.py
\end{verbatim}

    \item Download all required models:
\begin{verbatim}
python3 download_models.py <hf_token>
\end{verbatim}
\end{enumerate}

\subsection{Experiment workflow}

All figures can be reproduced using:

\begin{verbatim}
cd figures
bash run_figure.sh <num>
\end{verbatim}

Where \texttt{<num>} is one of:

\begin{verbatim}
1, 2, 7, 9, 10, 11, 12, 13, 14,
15, 17, rest
\end{verbatim}

The output figure will be generated at:

\begin{verbatim}
figures/figure-<num>/figure<num>.pdf
\end{verbatim}

Special cases:

\begin{itemize}
    \item \textbf{Figure 1:} Requires multiple vLLM versions using virtual environments.
    \item \textbf{Figure 10:} Requires two GPUs or two machines.
    \item \textbf{Figure 11:} Requires two different CPU systems.
    \item \textbf{Figure 13:} Compares RAM vs. SSD loading and requires cache clearing with sudo privileges.
\end{itemize}

\subsection{Evaluation and expected results}

Each script produces a PDF figure corresponding to the figures presented in the paper.

Due to hardware differences, results may exhibit small numerical variations. However, the overall trends, relative performance differences, and main conclusions should match those reported in the paper.

Successful reproduction is confirmed when:

\begin{itemize}
    \item Scripts complete without errors.
    \item Output figures are generated.
    \item Trends align with the published results.
\end{itemize}

\subsection{Methodology}

Submission, reviewing and badging methodology:

\begin{itemize}
  \item \url{https://www.acm.org/publications/policies/artifact-review-and-badging-current}
  \item \url{https://cTuning.org/ae}
\end{itemize}